\pdfoutput=1

\documentclass[11pt]{article}

\usepackage{acl}

\usepackage{times}
\usepackage{latexsym}

\usepackage[T1]{fontenc}

\usepackage[utf8]{inputenc}

\usepackage{microtype}

\usepackage{inconsolata}

\usepackage{graphicx}

\usepackage{amsmath,amsfonts,bm}

\def\eqref#1{equation~\ref{#1}}

\def\1{\bm{1}}

\DeclareMathAlphabet{\mathsfit}{\encodingdefault}{\sfdefault}{m}{sl}
\SetMathAlphabet{\mathsfit}{bold}{\encodingdefault}{\sfdefault}{bx}{n}

\newcommand{\smallsection}[1]{\noindent\textbf{#1}.\quad}

\DeclareMathOperator*{\argmin}{arg\,min}

\usepackage{hyperref}
\usepackage{url}
\usepackage{float}
\usepackage{hyperref}       
\usepackage{url}            
\usepackage{booktabs}       
\usepackage{amsfonts}       
\usepackage{nicefrac}       
\usepackage{microtype}      
\usepackage{xcolor}         
\usepackage{multirow,multicol}
\usepackage{graphicx}
\usepackage{enumitem}
\usepackage{listings}  
\usepackage{amsmath}
\usepackage{tcolorbox}
\usepackage{pdfpages} 
\usepackage{graphicx}   
\usepackage{subcaption} 
\usepackage{epstopdf}   
\usepackage{xcolor}
\usepackage{array}
\usepackage{tcolorbox}

\title{Next-Token Prediction Task Assumes Optimal Data Ordering for LLM Training in Proof Generation}

\author{Chenyang An$^1$\thanks{$\ $ Internship project at Microsoft Research.}, Shima Imani$^2$, Feng Yao$^1$, Chengyu Dong$^1$, Ali Abbasi$^3$,\\ \bf Harsh Shrivastava$^2$, Samuel Buss$^1$\thanks{$\ $  Corresponding authors.}, Jingbo Shang$^1$\footnotemark[2],  Gayathri Mahalingam$^2$\footnotemark[2],\\ \bf  Pramod Sharma$^2$\footnotemark[2], Maurice Diesendruck$^2$\footnotemark[2]\\University of California, San Diego$^1$ \quad
Microsoft Research$^2$ \quad Vanderbilt University$^3$\\
  \texttt{\{c5an, fengyao, cdong, sbuss, jshang\}@ucsd.edu}\\
  \texttt{\{shimanihs, hshrivastava, gmahalingam, pramod.sharma\}}@microsoft.com\\
  \texttt{ali.abbasi@vanderbilt.edu}\\
  \texttt{moglobal@gmail.com}
  }

\begin{document}
\maketitle
\begin{abstract}
In the field of large language model (LLM)-based proof generation, despite extensive training on large datasets such as ArXiv, LLMs still exhibit only modest performance on proving tasks of moderate difficulty. We believe that this is partly due to the widespread presence of suboptimal ordering within the data for each proof used in training.
For example, published proofs often follow a purely logical order, where each step logically proceeds from the previous steps based on the deductive rules. This order is designed to facilitate the verification of the proof's soundness, rather than to help people and models learn the discovery process of the proof. In proof generation, we argue that the optimal order for one training data sample occurs when the relevant intermediate supervision for a particular proof step in the proof is always positioned to the left of that proof step. We call such order the intuitively sequential order. We validate our claims using two tasks: intuitionistic propositional logic theorem-proving and digit multiplication. 
Our experiments verify the order effect and provide support for our explanations. We demonstrate that training is most effective when the proof is in the intuitively sequential order. Moreover, the order effect and the performance gap between models trained on different data orders can be substantial -- with an 11 percent improvement in proof success rate observed in the propositional logic theorem-proving task, between models trained on the optimal order compared to the worst order. Lastly, we define a common type of order issue in advanced math proofs and find that 17.3 percent of theorems with nontrivial proofs in the first two chapters of a widely used graduate-level mathematics textbook suffer from this issue. A detailed list of those proofs is provided in the appendix.
\end{abstract}

\section{Introduction}

Recent works have shown the potential of LLMs to perform mathematical reasoning and proof generation in both natural language and formalized environments 
\citep{yang2023leandojoa, welleck2021naturalproofs, lample2022hypertree, mikula2023magnushammer, wang2023dt}.

Meanwhile, leading models consume far more data than a human could (datasets like OpenWebMath contain 14B tokens), and still perform suboptimally on reasoning benchmarks 
\citep{paster2023openwebmath,azerbayev2023llemma, touvron2023llama,dubey2024llama, paster2023openwebmath}. 
This is also true for downstream reasoning tasks where LLMs need further fine-tuning \citep{yang2023leandojo, an2024learn}.
We identify this phenomenon as the training inefficiency problem.

One possible explanation for such low training efficiency in proof generation for LLMs is poor internal order within each training data sample in the math domain. For instance, consider the purely logical order that many textbook proofs follow, where each step logically proceeds from the previous steps based on the deductive rules. This order is widely adopted for presenting proofs publicly because it allows for straightforward verification of the proof’s correctness. However, from a proof discovery perspective, this order is not intuitive, i.e. not the order in which the proof is actually discovered. For human learners, the most intuitive order of proof steps is one in which all steps contributing to the discovery of a particular proof step are presented prior to it.  We call such order the intuitively sequential order. These steps, which aid in the discovery of a given proof step, are referred to as intermediate supervision for that step. 

\begin{figure*}[t]
    \centering    \includegraphics[width=0.9\linewidth]{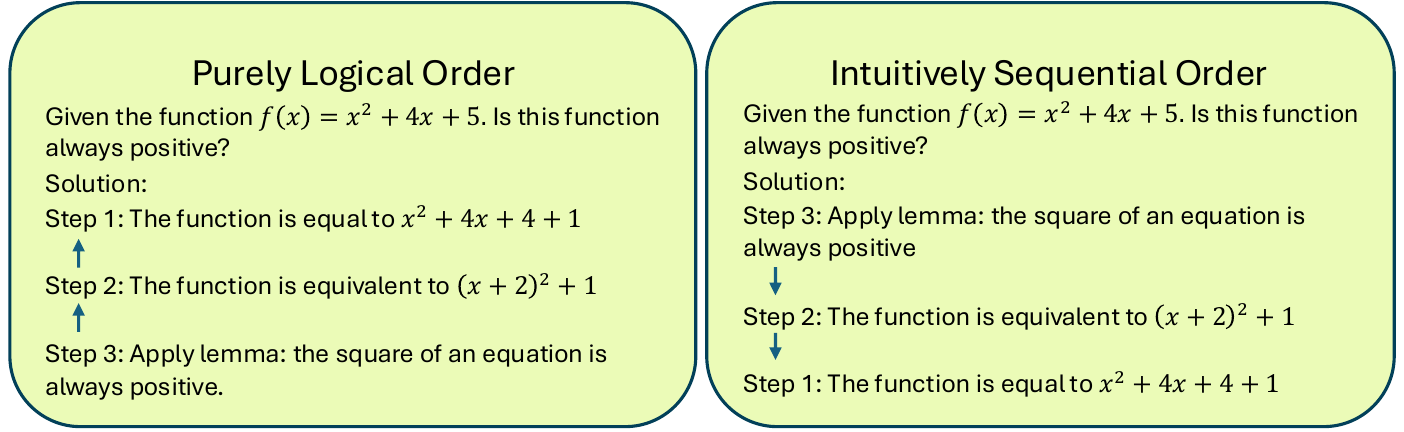}
    \caption{An illustrative example of the purely logical order versus the intuitively sequential order. The arrows indicate the direction of intermediate supervision.}   
\label{fig:illustrative-example}
\end{figure*}

Consider a concrete scenario where an intuitive sequential order is preferable to a purely logical one in learning and model training. When proving a theorem, a researcher quickly identifies a key lemma but must first perform preliminary steps to apply it. Textbooks typically present these steps first, following a purely logical order. However, in LLM training—and human learning—these steps are easier to derive after seeing the key lemma, which serves as intermediate supervision to these steps. Without the lemma, it is unclear why these preliminary steps are necessary. We refer to this issue as Delayed Insight Issue (DII)—a problem where crucial insights are introduced too late, disrupting intuition and understanding. To avoid DII, training data should present the lemma before the preliminary steps in such cases. DII represents one of the many causes of the difference between the purely logical order and the intuitively sequential order for advanced math proofs.

We give a simple illustrative example of DII in Figure~\ref{fig:illustrative-example}. In Figure~\ref{fig:illustrative-example}, Step 3 serves as the intermediate supervision that assists in predicting Step 2, which in turn serves as the intermediate supervision that assists in predicting Step 1. Inexperienced students typically struggle to understand this proof in the purely logical order on their first read, as it is unclear why the problem statement should lead to Step 1. If they are instead taught to first apply the lemma as in Step 3, the learning process becomes faster. In practice, the confusion of suboptimal order can be overcome through repeated reflection for humans. As for model training with next token prediction, without introducing more elaborate training objectives, the only way to achieve a similar effect is to modify the order of the training data. 

Given that (i) proofs and reasoning are often presented in the suboptimal order, and (ii) next-token prediction is the predominant method used for pre-training and fine-tuning cutting-edge autoregressive LLMs, we hypothesize that the training inefficiency problem of LLMs in solving reasoning tasks is partly due to the suboptimal order of each proof's data during training. We claim that the intuitively sequential order is the optimal order for fine-tuning models. We show that there are two causes for the order effect. Assume we have proof step $a$, and proof steps $\{a_i\}$ that serve as intermediate supervision toward $a$, and they occur in the order $``a, \{a_i\}"$. The first cause is that when learning to predict $a$, the model will not directly learn to use the supervision $\{a_i\}$, because it does not look forward during training. The second cause is that this order also induces the model to fit the non-existent dependency of $\{a_i\}$ on $a$, which we call the spurious relation. Note that in reasoning tasks, intermediate supervision is usually another sequence of reasoning steps.

To verify our claims, we conduct experiments on two downstream tasks: (1) 4-by-4 digit multiplication and, (2) intuitionistic propositional logic theorem-proving. Various metrics were employed to measure model performance. The results consistently show that models fine-tuned/trained-from-scratch on the intuitively sequential proofs outperformed those fine-tuned/trained-from-scratch on proofs in other orders. 

Our experiments also validate the two causes for the order effect we propose. We find that models do not learn to use future tokens to predict the current proof step. 
We also show evidence in the multiplication task that models are negatively impacted by spurious step dependencies learned during training when it's trained on data in suboptimal orders. 

To demonstrate the prevalence of ordering issues in mathematical texts, we analyze the first two chapters of a widely used graduate-level textbook in analysis, which contain a total of 57 theorems with nontrivial proofs. We find that 17.3 percent of these proofs exhibit DII. A detailed list of the affected proofs is provided in Appendix~\ref{example:folland}.

Our main contributions are summarized as follows:
\begin{itemize}[noitemsep, topsep=0pt]
    \item We find that data order significantly affects LLM training efficiency on reasoning tasks, which we refer to as the order effect.  
    \item We claim optimal order to be the intuitively sequential order, where intermediate supervision is placed on the left.
    \item Through the propositional logic theorem-proving task and the digit multiplication task,
    our experiments show that this order effect is due to next-token prediction. Two causes are (1) LLMs do not look forward during training, (2) LLMs may learn spurious token dependencies if trained with data samples in suboptimal order.
    \item We show that the Delayed Insight Issue—a specific ordering problem—frequently arises in advanced mathematical proofs, as illustrated by 9 examples from a popular textbook listed in the appendix.
\end{itemize}

\section{Related Work}
\smallsection{Problem of next-token prediction training}
Existing research has exposed weaknesses in the next-token prediction mechanism. For example, \cite{berglund2023reversal} showed the Reversal Curse: a model trained on ``A is B" does not know that ``B is A". \cite{golovneva2024reverse} proposed to train a model both sequentially and reversely, using two tasks together to enhance performance. In contrast, our work demonstrates that optimal orders exist for next-token prediction learning, specifically by the rule of positioning intermediate supervision to the left. 
\cite{bachmann2024pitfalls} highlighted a similar toy problem known as the star-path problem, where initial errors substantially reduced the model's ability to predict later tokens.

\smallsection{Order effect on LLM: context}
Several works have studied how disrupting order in prompts might affect model performance. \cite{chen2024premise} showed that LLMs achieve the best performance on GSM8K when the premise order aligns with the
context required in intermediate reasoning steps. \cite{liu2024lost} discover the lost in-the-middle phenomenon in the long-context scenario: LLM performance is best when relevant information is placed at the beginning or the end of the input context, while the performance is worst when the LLM needs to utilize information in the middle.

\smallsection{Order effect on LLM: digits}
For long-digit addition, prior works demonstrated that reversing numbers in training data can help LLMs achieve better performance on digit-addition tasks \citep{lee2023teaching, zhou2023algorithms, zhou2024transformers}. While their work only applies to digit computation since they employed digit-wise reversal, our work studies the positional role of intermediate supervision in general for the proof generation task.
\section{Preliminary}
\subsection{Problem setting}
Given a nondeterministic algorithm $\mathbf{A}$ and input $r_0$ from the problem space, we assume:
\begin{enumerate}
    \item \smallsection{Finiteness} Algorithm $\mathbf{A}$ generates a finite sequence of steps $\mathbf{A}(r_0)=r_1, \mathbf{A}(\{r_0,r_1\})=r_2,...,\mathbf{A}(\{r_i\}_{i=0}^{n-1})=r_{n_{r_0}}$, where $r_{n_{r_0}}$ is the last step that $\mathbf{A}$ will generate when given $r_0$, which usually is the final answer. Here $n_{r_0}$ is the position of the last token (final answer) generated by algorithm $\mathbf{A}$ with input $r_0$.
    \item \smallsection{Intuitively sequential order} Each step $r_i$ highly depends on $r_j$ for $1<=j<i$, i.e. the probability for $\mathbf{A}(\{r_i\}_{i=0}^{j-1})=r_j$ is high for all  $j$. For every $r_0$, $\{r_i\}_{i=0}^m$ serves as intermediate supervision for models to predict $r_{m+1}$ for all $0<=m<n_{r_0}$. We call this order the intuitively sequential order. All other orders of the sequence have a weaker right-to-left dependency than the aforementioned order. 
\end{enumerate}

Given a model $M_\theta$ with $\theta$ as weights of the model, our goal is to enable $M_\theta$ to imitate algorithm $\mathbf{A}$ through the next-token prediction task. Specifically, we aim to find weights $\theta_0$ such that $\theta_0 = \argmin_\theta \frac{1}{n_{r_{0}}}\Sigma_{m=1}^{n_{r_0}}\mathbb{E}_{r_0}|M_\theta(\{r_i\}_{i=1}^{m})-\mathbf{A}(\{r_i\}_{i=1}^{m})|$. 
To put it in another way, we want the model to predict the entire sequence $\{r_i\}_{i=1}^{n_{r_0}}$ given $r_0$.  Importantly, the loss function does not exclusively focus on $r_{n_{r_0}}$ but also accounts for the intermediate steps $r_{m}$ for $m=1,...,r_{n_{r_0-1}}$ when computing loss. This scenario arises in practice in theorem-proving tasks, where we are interested in both the intermediate proofs and the final answer.

\subsection{Causes of the order effect}
In preparation for training model $M_\theta$ to learn $\mathbf{A}$, we collect a set of training data. Each data sample is a finite sequence of steps $\{r_i\}_{i=0}^{n_{r_{0}}}$. In the next-token prediction task, given a training data point $r=\{r_i\}_{i=0}^{n_{r_0}}$, autoregressive LLMs update their parameter by minimizing the loss of $\Sigma_{m=1}^{n_{r_0}}|M_\theta(\{r_i\}_{i=0}^m)-\mathbf{A}(\{r_i\}_{i=0}^m)|$ over batches. 
Now, without loss of generality, if the data point $\{r_i\}_{i=0}^{n_{r_{0}}}$ in the set of training data satisfies the assumption above, and is reordered into $\{r_0,r_{n_{r_0}},r_1,...,r_{n_{r_0}-1}\}$, the following things happen:
\begin{itemize}
    \item \smallsection{Model cannot look forward: intermediate supervision fails to help if placed on the right} The model will not be able to learn the dependency of $r_{n_{r_0}}$ on $\{r_1,...,r_{n_{r_0}-1}\}$. Through the next-token prediction training loss, given $r_i\in \mathbf{r}=\{r_i\}_{i=0}^{n_{r_{0}}}$, the model has no way to look forward and use $r_j$ with $j>i$ to help learn to predict $r_i$. The model would find it much harder to learn to predict $r_{n_{r_0}}$ directly from $r_0$, since this task is not split into sub-tasks like given $\{r_0\}$ predict $r_1$, given $\{r_0, r_1\}$ predict $r_2$, and so on \citep{wies2022sub}.

    \item \smallsection{Model may learn spurious relations} Since $r_{n_{r_0}}$ is positioned before $\{r_1,...,r_{n_{r_0}-1}\}$, the model might learn a spurious dependency of $\{r_1,...,r_{n_{r_0}-1}\}$ on $r_{n_{r_0}}$ at the early stage of training, and could take many training steps to realize that $r_{n_{r_0}}$  contains no information for predicting $\{r_1,...,r_{n_{r_0}-1}\}$.

\end{itemize}
Later in Sec~\ref{sec:result} we display evidence supporting our claim that the aforementioned two reasons are among the causes of the order effect.
\section{Experimental Setup}
\subsection{Tasks and datasets}
\subsubsection{Intuitionistic propositional logic theorem-proving}
For the first dataset, we use theorem statements and proofs in intuitionistic propositional logic (PropL), as introduced in \cite{an2024learn}. A proof consists of tactics and intermediate states, where a tactic refers to an actual proof step, and a state represents what remains to be proved after a tactic is applied. Both the statements and the proofs are written in Lean, a popular formalized environment for theorem-proving in the mathematical community \citep{de2015lean}. We have two types of PropL data. In the first type of data, all proofs follow the original sequential order (referred to as SEQ order, see Table~\ref{Table:datatype}). We note that SEQ order follows the intuitively sequential order: all the previous tactics and states help generate the next tactic and state. For the second type, the proofs are reversed (referred to as SER order), such that the last tactic and the last state become the first tactic and the first state, the second-to-last tactic and the second-to-last state become the second tactic and the second state, and so forth. Note that this dataset was published after the Llama-2 family was released so the model hasn't seen this dataset during pre-training. See an example of a statement and a proof in Appendix~\ref{appendix:data-example}.

For training and testing, we select all statements in the PropL dataset that are proved using 3, 4, or 5 tactics, totaling 21,000 instances. The last 1,000 statements are reserved for testing so that we have 20,000 training instances and 1,000 testing instances. During fine-tuning, the statement together with its proof (in SEQ or SER order) forms one training data sample. Models see each training data sample at most once during training.

\subsubsection{4-by-4 digit multiplication}

For our second dataset, we choose the 4-digit by 4-digit multiplication task as a proof generation task. After a pair of 4-digit numbers (num1, num2) is picked, the problem is generated as: ``Tell me what is num1*num2 and prove it". A proof of a problem consists of the steps where we induce different axioms of the integer ring to derive the middle steps for computation, and the last step of the proof is the final answer of the multiplication. 

We use an algorithm to generate the problems, the proofs, and the final answers of 4-digit by 4-digit multiplication. It has been shown that such problems, even prompted with ``Let's think step by step'', are challenging for cutting-edge models like GPT-4 \citep{liu2023goat}. We create three types of proofs with the same proof steps but in different orders, whose compositions are summarized in the table. Examples of all three types of data can be found in the Appendix~\ref{appendix:data-example}. Note that for the multiplication task, for proofs in the SEQ order, given a proof step, all previous steps serve as the natural intermediate supervision toward that step, hence making this dataset ideal for testing our claims.

For training and testing, we first enumerated all 4-digit by 4-digit tuples. The list of tuples was then randomly shuffled. We used the first 1,000,000 tuples as our training number tuples. Together with the proofs, they formed our training dataset. We used the last 1,000 tuples to construct our testing number tuples. 
\begin{table*}[t]
\centering
\caption{Three Types of Data. See Appendix~\ref{appendix:data-example} for examples.}
\label{Table:datatype}
\begin{tabular}{|c|c|}
\hline
\textbf{Type Name} & \textbf{Proof Steps Composition} \\
\hline
Sequential (SEQ)  & Problem + Sequential Proof + Final Answer \\
\hline
Partially Sequentially Reversed (PSER) & Problem + Sequentially Reversed Proof + Final Answer  \\
\hline
Sequentially Reversed (SER) & Problem +  Final Answer + Sequentially Reversed Proof \\
\hline
\end{tabular}
\end{table*}

\subsection{Metrics}
\smallsection{Metrics for the intuitionistic propositional logic theorem-proving task} 
For the theorem-proving task, we evaluate model performance using two key metrics. The first metric is tactic prediction correctness, which assesses whether the tactics generated by the model can prove the theorem. The second metric is proof generation correctness, which not only checks if the generated tactics prove the theorem, but also ensures that all intermediate states are correctly generated. The entire evaluation process is automated using Lean. We input the theorem statement and the model-generated tactics into Lean. Lean computes a state after applying the given tactic to the given state. We then compare the model-generated intermediate states with the states computed by Lean. Lean automatically determines whether the tactics successfully prove the theorem.

\smallsection{Metrics for the multiplication task}
In contrast to prior works, we treat 4-digit by 4-digit computation as a proof generation task, not as a computation task \citep{liu2023goat, zhou2023algorithms, zhou2024transformers, lee2023teaching}. We evaluate model performance using the proof correctness metric, final answer correctness metric, and Step 1 correctness metric, where Step 1 refers to the first step in the proof according to the optimal order in the multiplication task. For the proof correctness metric, our model is prompted with only the problem, ``Tell me what num1*num2 is and prove it'', and a successful generation of the model requires every proof step in the generation to be correct, together with the final answer. It is in parallel with the usual proof generation scenario, where one is given a statement and is asked to generate the proof to prove whether the statement is true or false. In mathematical reasoning, simply giving out a true or false answer is not enough without correct convincing arguments. For the final answer prediction task, we prompt the model in two ways. The first prompt contains only the problem. The second prompt contains both the problem and all the intermediate proof steps except the final answer correspondingly. For both prompts, we check the correctness of the final answer. For Step 1 correctness metric, we prompt the model with the problem and the intermediate proof steps, except for Step 1. We then check the correctness of Step 1 output by the model. 

\subsection{Models and hyperparameters}
We use Gemma-2B \citep{team2024gemma} and Llama-2-7B-hf \citep{touvron2023llama} as our proof generators and the default pre-trained tokenizers for both of the models, even in train-from-scratch experiments. We use the function AutoModelForCausalLM.from\_config from Huggingface to randomly initialize the models. Unless otherwise stated, the models that we fine-tune are pre-trained with default weights. The learning rate of all training processes is set to be $5\times 10^{-5}$. We use Adam as our optimizer \citep{diederik2014adam}.  All training processes are performed on 4 NVIDIA A100, with the per-device batch size set to 1. This makes the batch size to be 4 for all experiments. All experiments are completed within 24 hours, except for the train-from-scratch experiments, which take about 3 days to finish. Every training process is based on prediction with cross-entropy loss.

During the experiments, for the multiplication task, we fine-tune Gemma-2b and Llama-2-7b-hf on SEQ, PSER, and SER datasets, resulting in a total of six models. We also train from scratch randomly initialized Gemma-2b on SEQ, PSER, and SER as well so that we have additional three models. We call models that are fine-tuned/trained from scratch on SEQ, PSER, and SER datasets as SEQ models (3 in total), PSER models (3 in total), and SER models (3 in total) respectively. 

For the theorem-proving task, we fine-tune Llama-2-7b-hf on SEQ and SER datasets. We call the two models the SEQ model and the SER model.

\begin{figure*}[t]
    \centering
    \begin{subfigure}{0.31\linewidth}
        \centering
        \includegraphics[width=\linewidth]{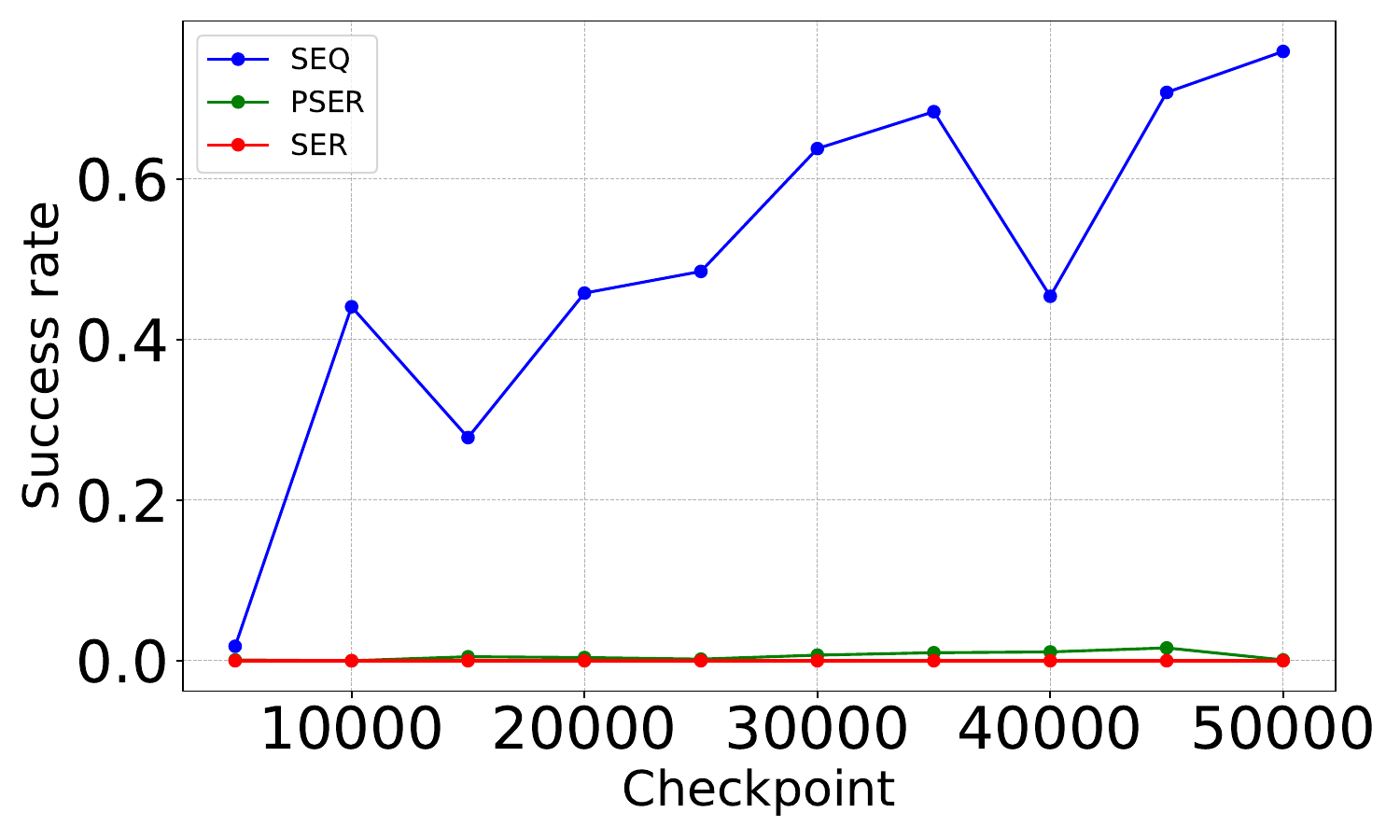}
        \caption{Pre-trained Gemma-2b tested on proof generation}
        \label{fig:gemma-proof-generation}
    \end{subfigure}
    \hfill
    \begin{subfigure}{0.31\linewidth}
        \centering
        \includegraphics[width=\linewidth]{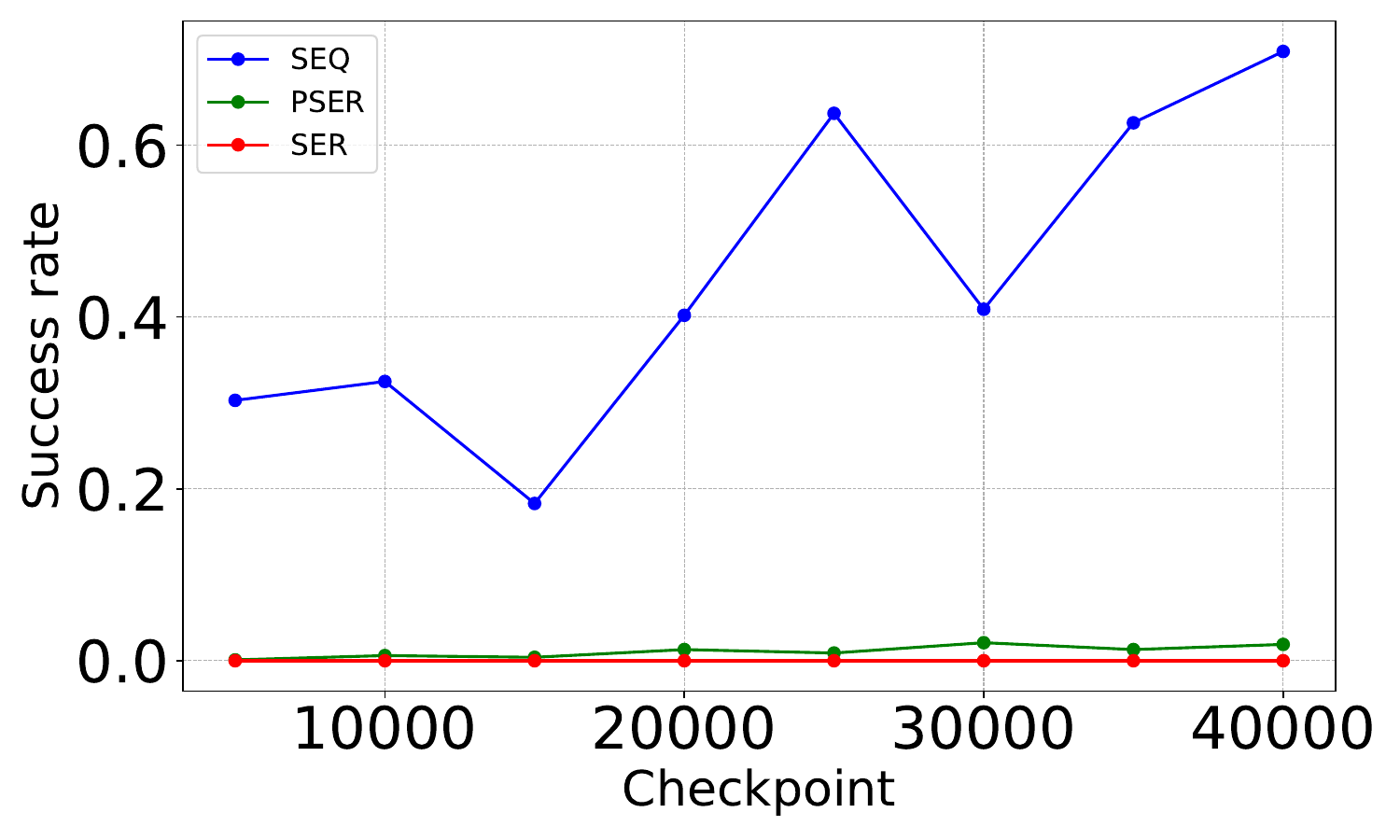}
        \caption{Pre-trained Llama-7b tested on proof generation}
        \label{fig:llama-proof-generation}
    \end{subfigure}
    \hfill
    \begin{subfigure}{0.31\textwidth}
        \centering
        \includegraphics[width=\linewidth]{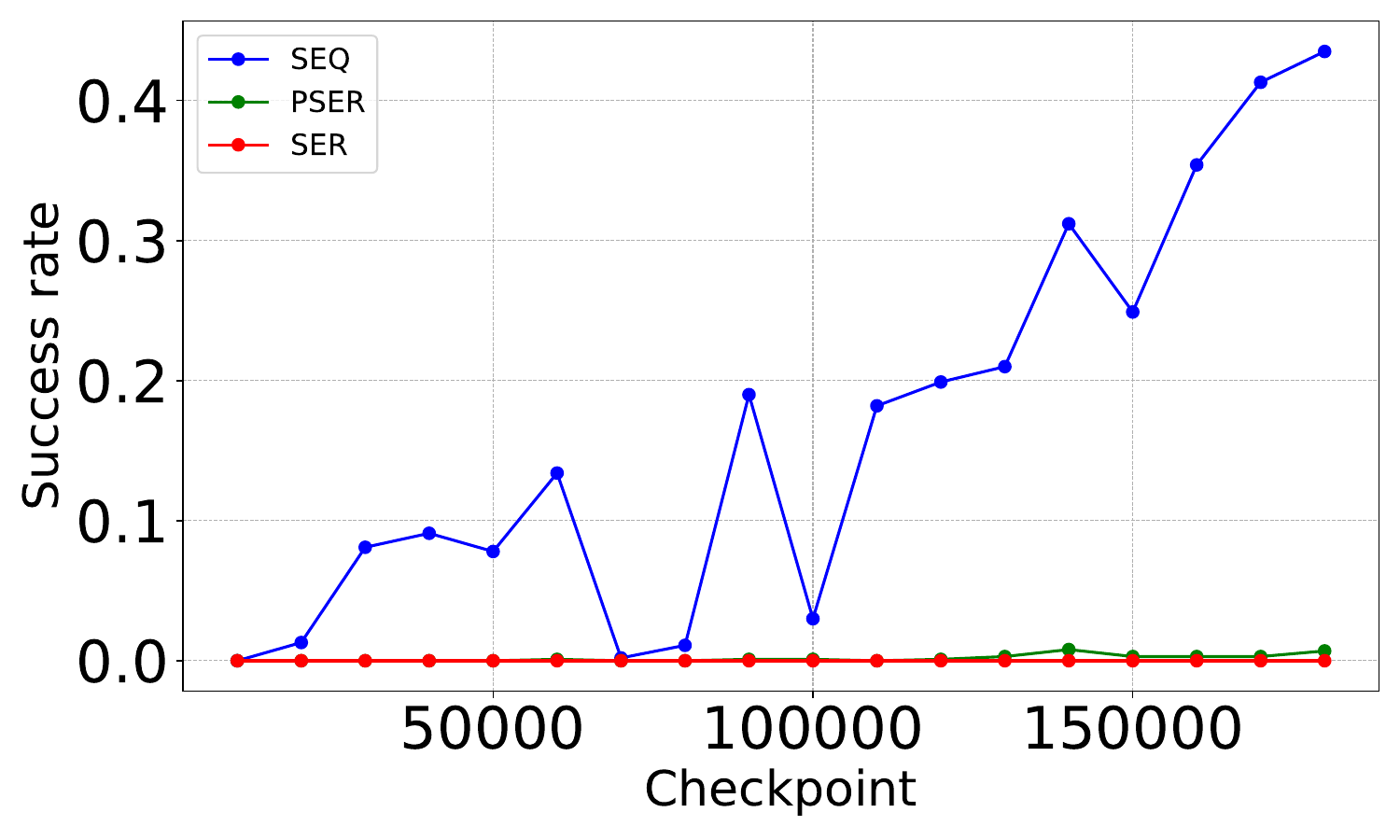}
        \caption{Train-from-scratch Gemma-2b tested on proof generation}
        \label{fig:gemma-train-from-scratch-proof-generation}
    \end{subfigure}
    \hfill
    
    \vfill
    \begin{subfigure}{0.31\linewidth}
        \centering
        \includegraphics[width=\linewidth]{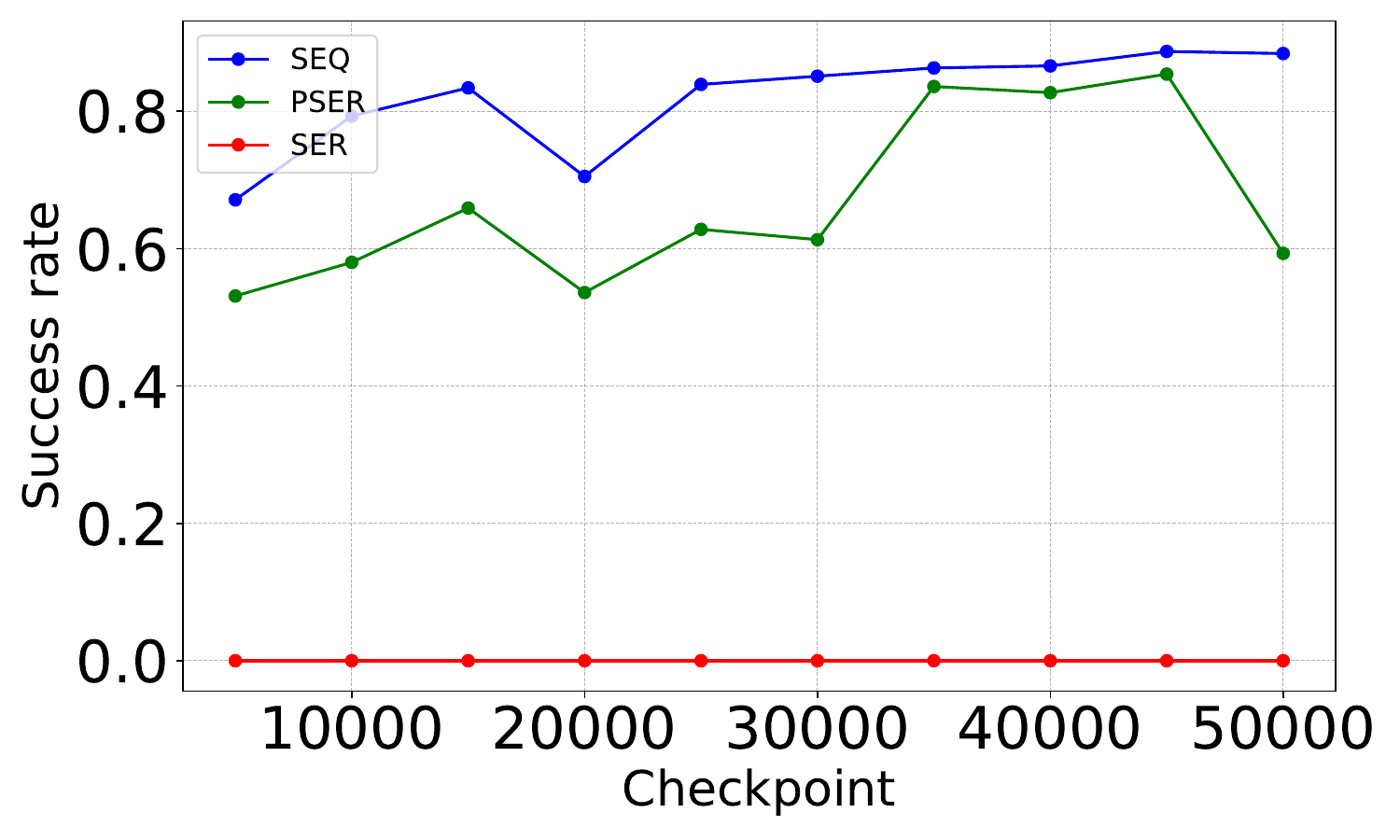}
        \caption{Pre-trained Gemma-2b tested on final answer}
        \label{fig:gemma-final-answer}
    \end{subfigure}
    \hfill
    \begin{subfigure}{0.31\linewidth}
        \centering
        \includegraphics[width=\linewidth]{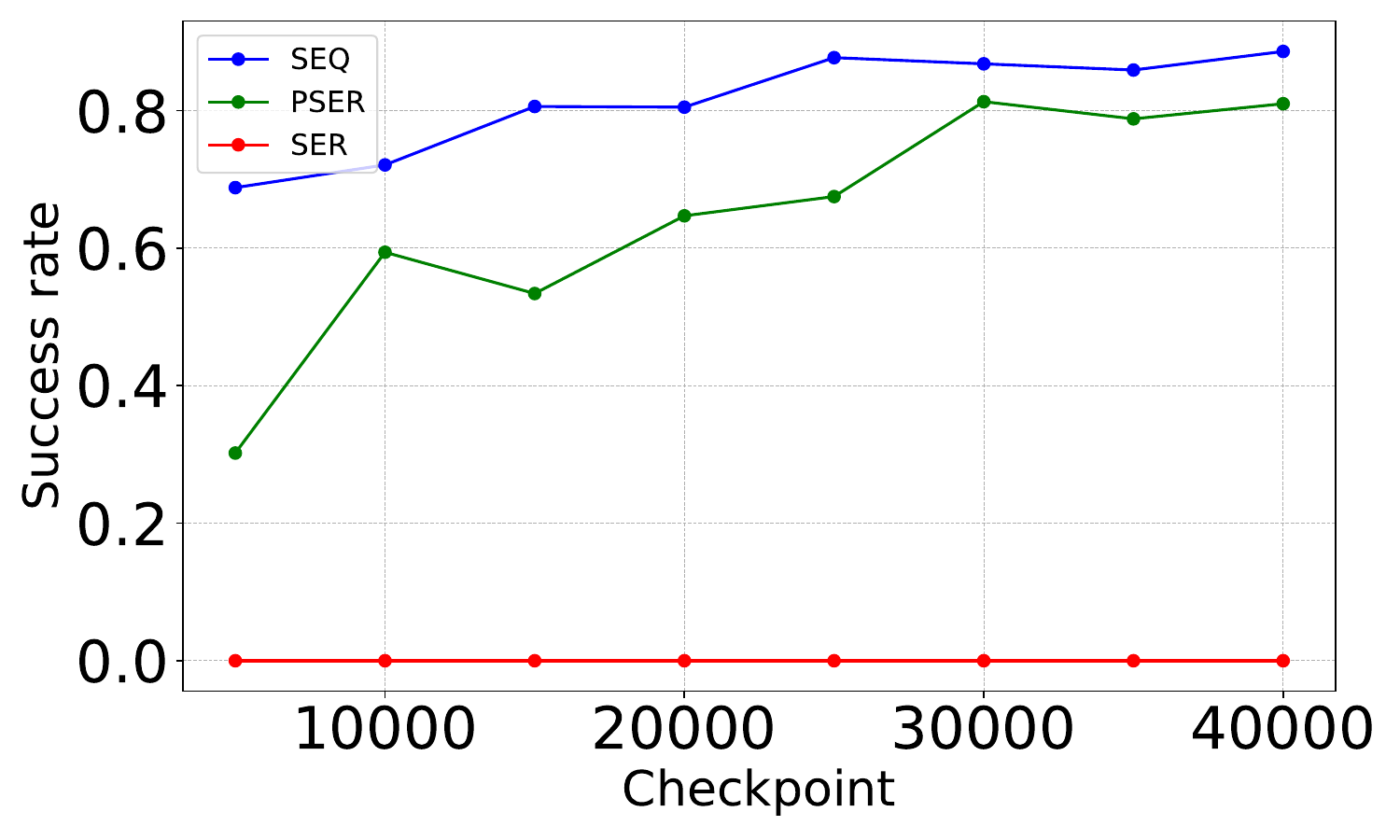}
        \caption{Pre-trained Llama-7b tested on final answer}
        \label{fig:llama-final-answer}
    \end{subfigure}
    \hfill
    \begin{subfigure}{0.31\linewidth}
        \centering
        \includegraphics[width=\linewidth]{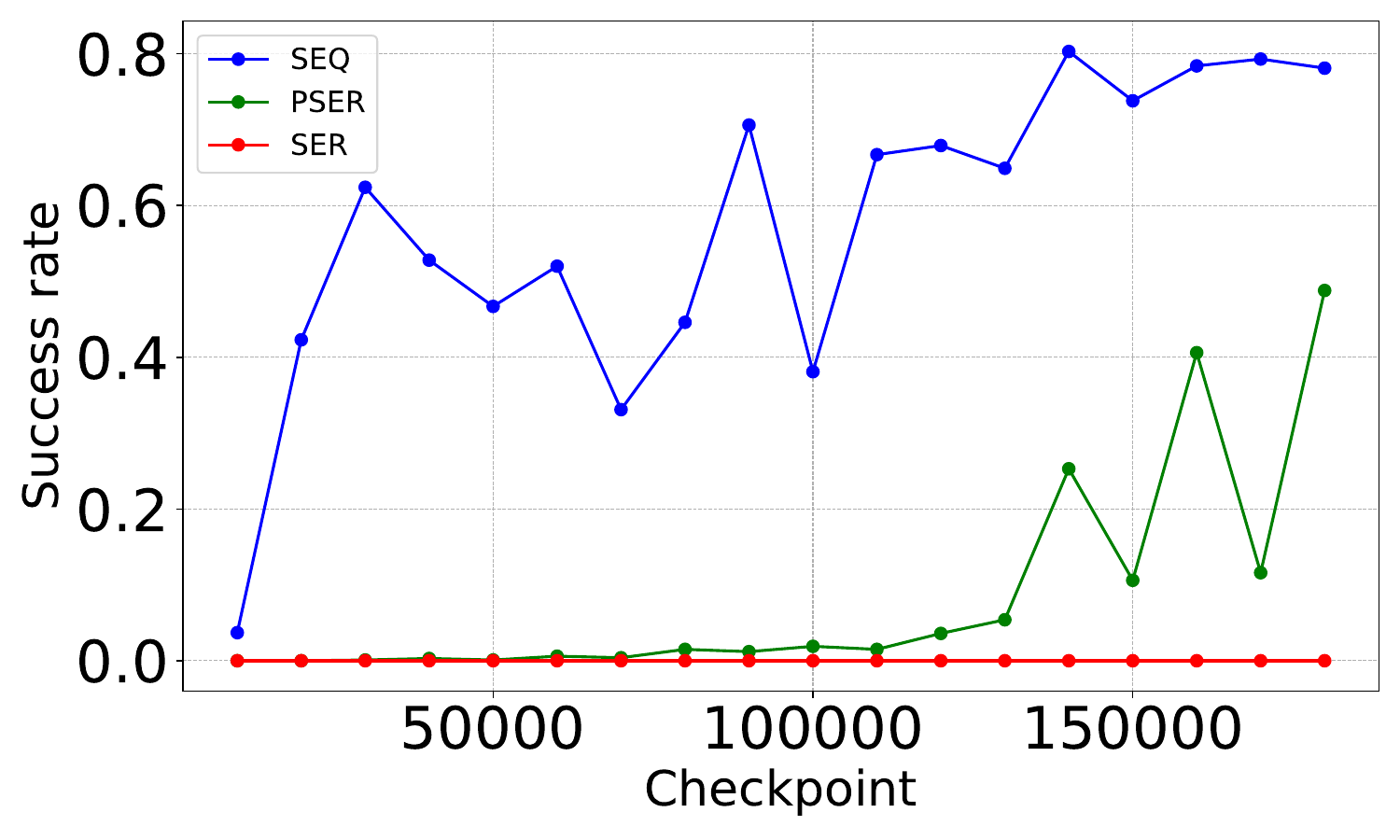}
        \caption{Train-from-scratch Gemma-2b tested on final answer }
        \label{fig:gemma-train-from-scratch-final-answer}
    \end{subfigure}
    \hfill
    \caption{Experimental results on the multiplication task measured by the metrics of proof generation correctness and final answer correctness. SER models achieves 0 percent success rate for all cases. In (a), (b) and (c), models are prompted with the problems only and performances are measured by the proof correctness metric. In (d), (e) and (f), SEQ and PSER models are prompted with both the problems and the intermediate steps, whereas SER model is prompted with the problems only. Performances are measured by the final answer correctness metric. Results show no evidence for the claim that the model can look forward during training.}

\end{figure*}
\label{sec:result}
\smallsection{Zero-shot performance of the models on the multiplication and theorem-proving tasks}
We begin by evaluating the zero-shot performance of both Gemma-2b and Llama-2-7b-hf on two tasks with 3-shots examples. For the theorem-proving task, we prompt Llama-2-7b-hf with a theorem statement and request it to generate a proof in Lean, see Figure~\ref{fig:prompt}f. The model, however, fails to produce any correct solution measured by either the tactic prediction metric or the proof correctness metric, achieving a success rate of 0 out of 1000 test cases.
For the multiplication task, we conduct tests using 100 cases. The models are provided with two types of prompts. The first prompt is, ``Tell me what num1 * num2 is and prove it." The second prompt incorporates a chain-of-thought (CoT) instruction: ``Tell me what num1 * num2 is and prove it. Let's think step by step." Both two prompts also include 3 examples in their corresponding formats. Despite these variations, both pre-trained Gemma-2b and Llama-2-7b-hf fail on the 4-by-4 multiplication task with a success rate of 0/1000 based on proof correctness and final answer accuracy. This result aligns with expectations, as the 4-by-4 multiplication task poses significant challenges even for more advanced models like GPT-4 \citep{liu2023goat}.

\begin{figure*}[t]
    \centering
    \begin{subfigure}{0.45\linewidth}
        \centering
        \includegraphics[width=\linewidth]{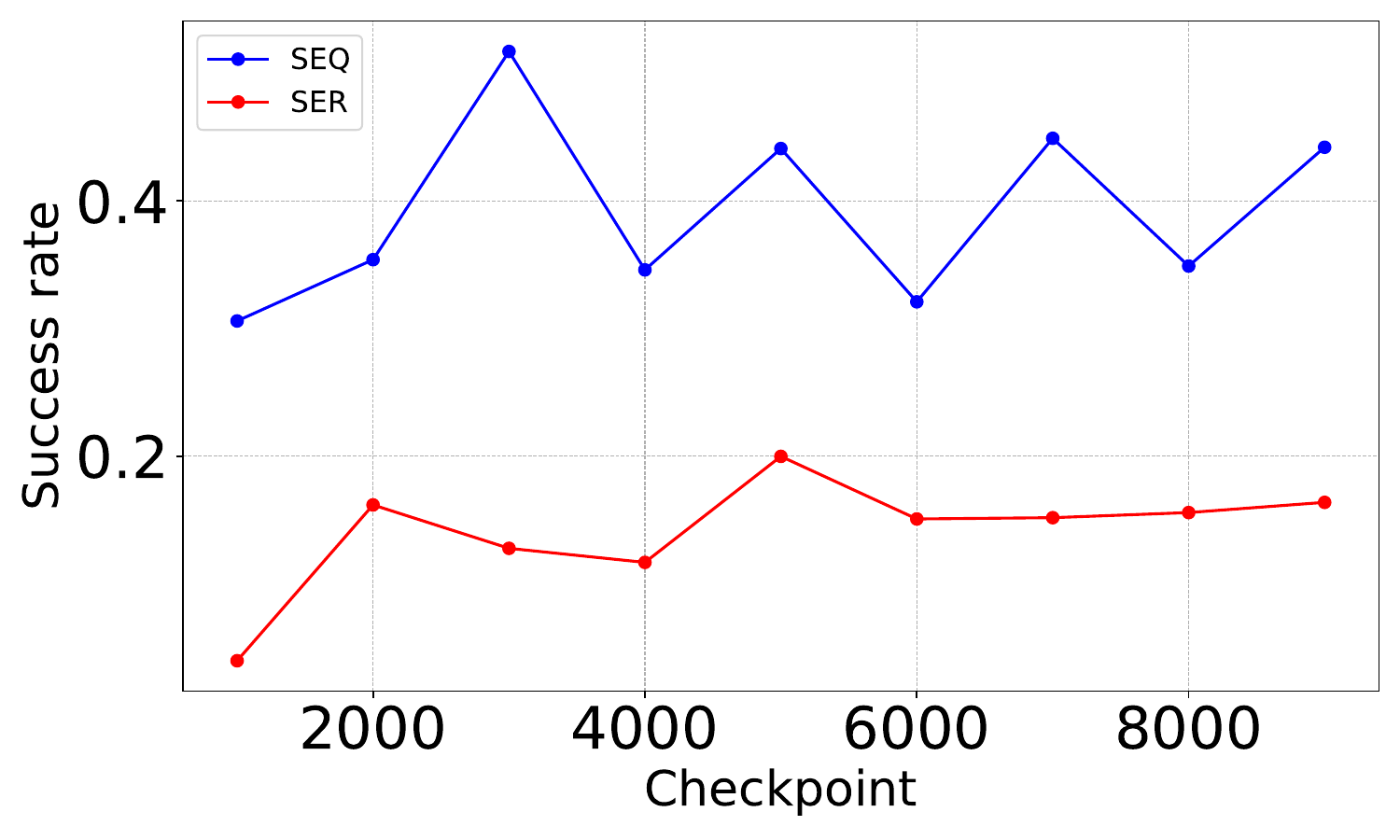}
        \caption{Pre-trained Llama-2-7b-hf tested on tactic prediction}
        \label{fig:propl-llama-tactic}
    \end{subfigure}
    \hfill
    \begin{subfigure}{0.45\linewidth}
        \centering
        \includegraphics[width=\linewidth]{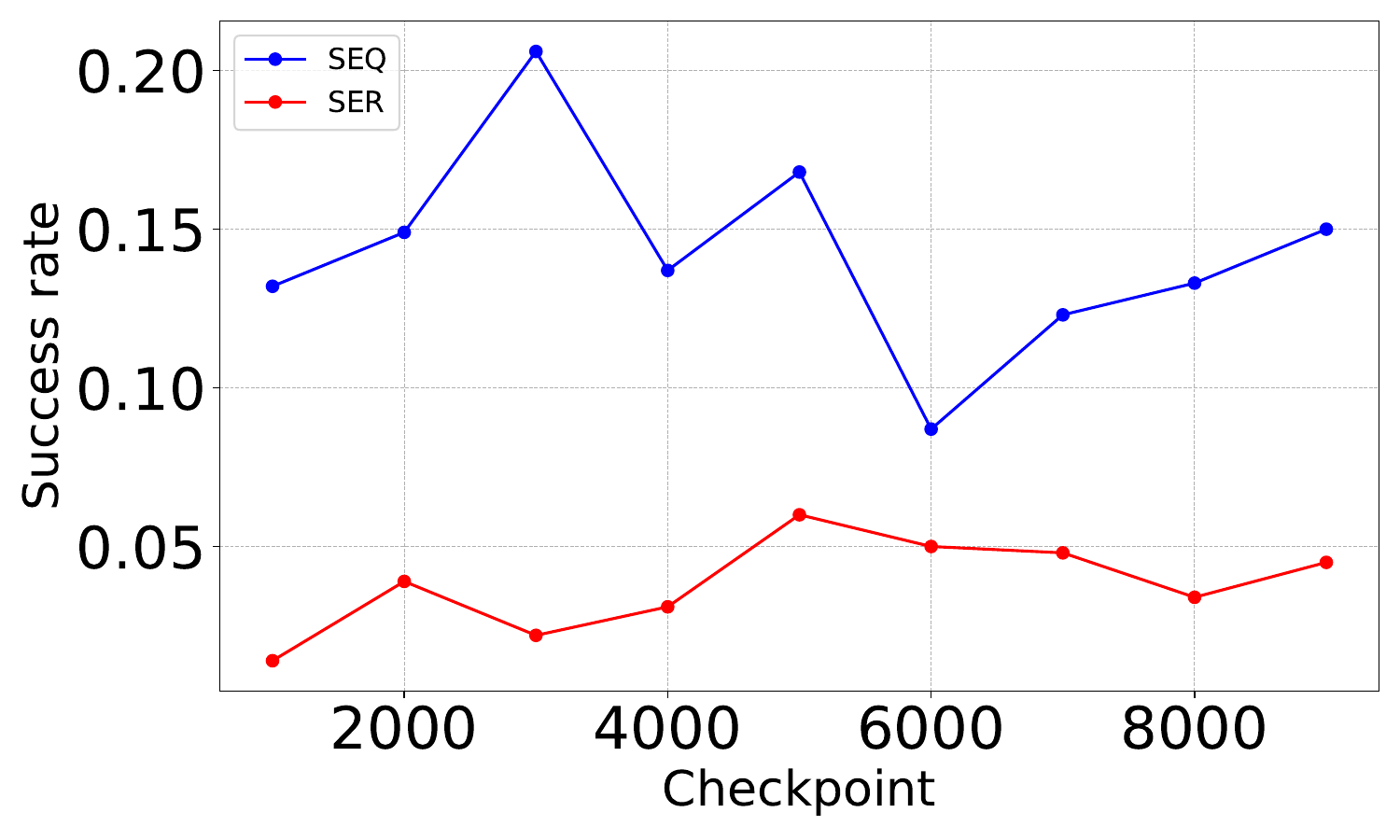}
        \caption{Pre-trained Llama-2-7b-hf tested on proof generation}
        \label{fig:propl-llama-proof}
    \end{subfigure}
    \caption{Experiment result on the intuitionistic propositional logic theorem-proving task. In (a) and (b), models are prompted with the theorem statement. For (a), performance is measured by the tactic prediction metric. For (b), performance is measure by the proof generation metric. With the same training steps, the model fined-tuned on SEQ proofs perform better than the model fined-tuned on SER proofs}
    \label{fig:result}
\end{figure*}

\smallsection{Main observation: training efficiency is maximized when models are trained on intuitive sequential order}
We conduct experiments on both the intuitionistic propositional logic theorem-proving task and the multiplication task.
For all the nine models that are fine-tuned/trained from scratch for the multiplication task, we prompt the models with ``Tell me what is num1*num2 and prove it'', where (num1, num2) is in the testing data. See Figure~\ref{fig:prompt}a,~\ref{fig:prompt}b,~\ref{fig:prompt}c for prompting configuration. 
We then measure the generated output of the model by the proof correctness metric. 
For the intuitionistic propositional logic theorem-proving task, we prompt the SEQ and SER models with the theorem statement. See Figure~\ref{fig:prompt}f for prompting configuration. We use both the tactic prediction metric and the proof correctness metric to measure model performance for this task. As shown in Figure~\ref{fig:gemma-proof-generation},~\ref{fig:llama-proof-generation},~\ref{fig:gemma-train-from-scratch-proof-generation},~\ref{fig:propl-llama-tactic},~\ref{fig:propl-llama-proof}, SEQ models learn the task a lot more efficiently than the models PSER models and SER models for the multiplication task. For the propositional logic theorem-proving task, the SEQ model also outperforms the SER model by 20 percent measured by the tactic prediction metric. Note that for the multiplication task, for all three base models, PSER models do get a small number of test problems right in the second half of the training, and SER models failed all test data consistently. This performance difference is in line with our claim that putting more intermediate supervision on the left will improve model training efficiency in reasoning. 

\smallsection{LLMs don't look forward in training}
For this experiment, we focus on the multiplication task. We use the same 9 models as in our first experiments, but for SEQ and PSER models, we prompt the model with both the problems and the intermediate steps excluding the final answer, so that we exclude the snow-balling effect (the effect that the earlier errors model makes will affect model performance in later outputs). See Figure~\ref{fig:prompt}d,~\ref{fig:prompt}e for prompting configuration. For SER models, since there is no intermediate step before the final answer in the data, we prompt it with the problem only. See Figure~\ref{fig:prompt}c for prompting configuration in this case. We use final answer prediction correctness for this experiment to measure model performance. As displayed in Figure~\ref{fig:gemma-final-answer},~\ref{fig:llama-final-answer},~\ref{fig:gemma-train-from-scratch-final-answer}, in this setting, for all three base models, models that are fine-tuned/trained from scratch on SER got 0 final answers correct consistently. The results of SER show that LLMs don't look forward during training, i.e. fail to learn to use intermediate supervision on the right in training. 

We also note that when both the problem and all intermediate proof steps (except for the final answer) are provided as prompts to the models, SEQ models still outperform PSER models in terms of final answer correctness, as shown in Figure~\ref{fig:gemma-final-answer},~\ref{fig:llama-final-answer},~\ref{fig:gemma-train-from-scratch-final-answer}. See Figure~\ref{fig:prompt}d,~\ref{fig:prompt}e for prompting configuration. 
By giving the correct intermediate steps as prompts to both SEQ and PSER models, we eliminate the snowballing effect during model generation.
\begin{figure*}[t]
    \centering    
    \begin{subfigure}{0.31\linewidth}
        \centering
        \includegraphics[width=\linewidth]{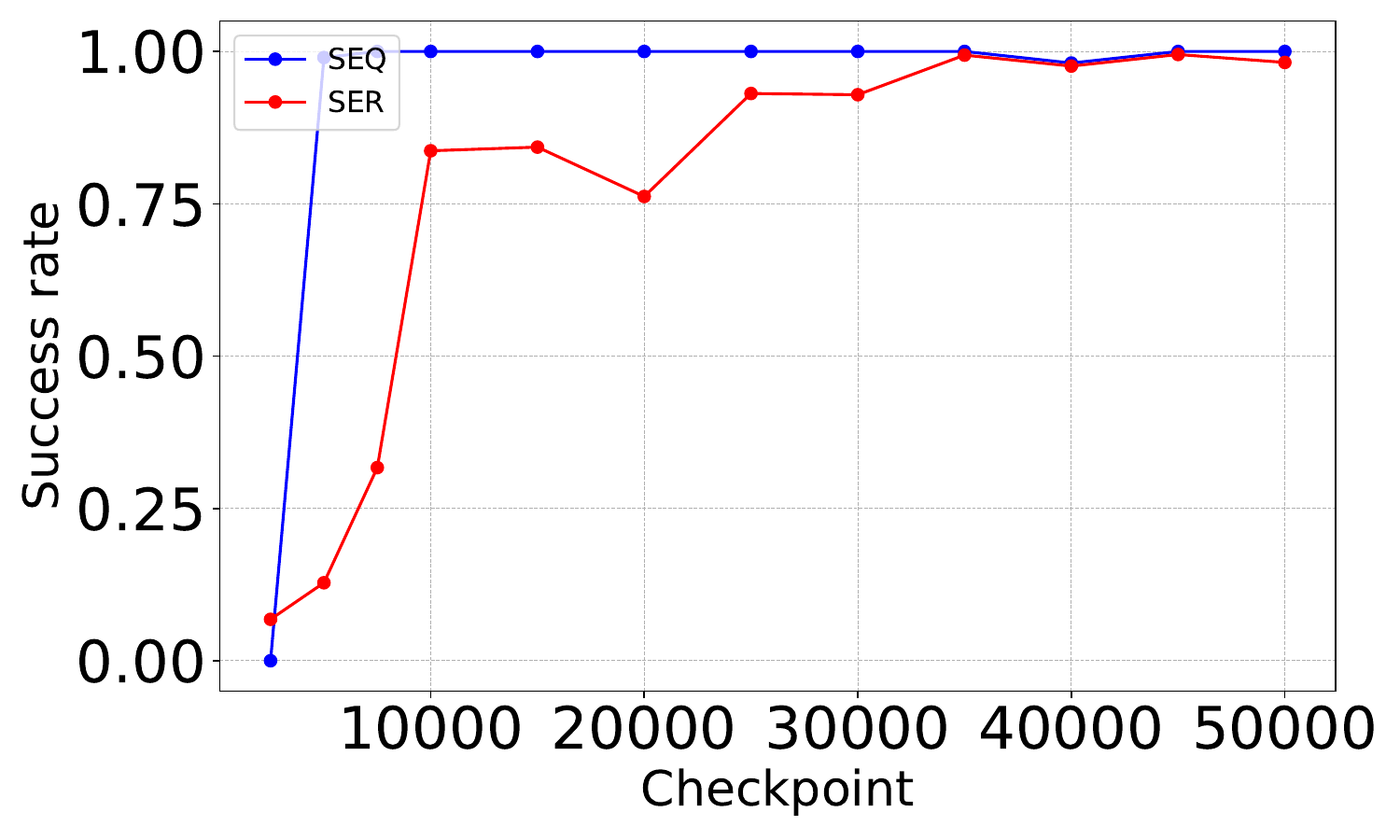}
        \caption{Pre-trained Gemma-2b tested on Step 1}
        \label{fig:gemmea-check-step}
    \end{subfigure}
    \hfill
    \begin{subfigure}{0.31\linewidth}
        \centering
        \includegraphics[width=\linewidth]{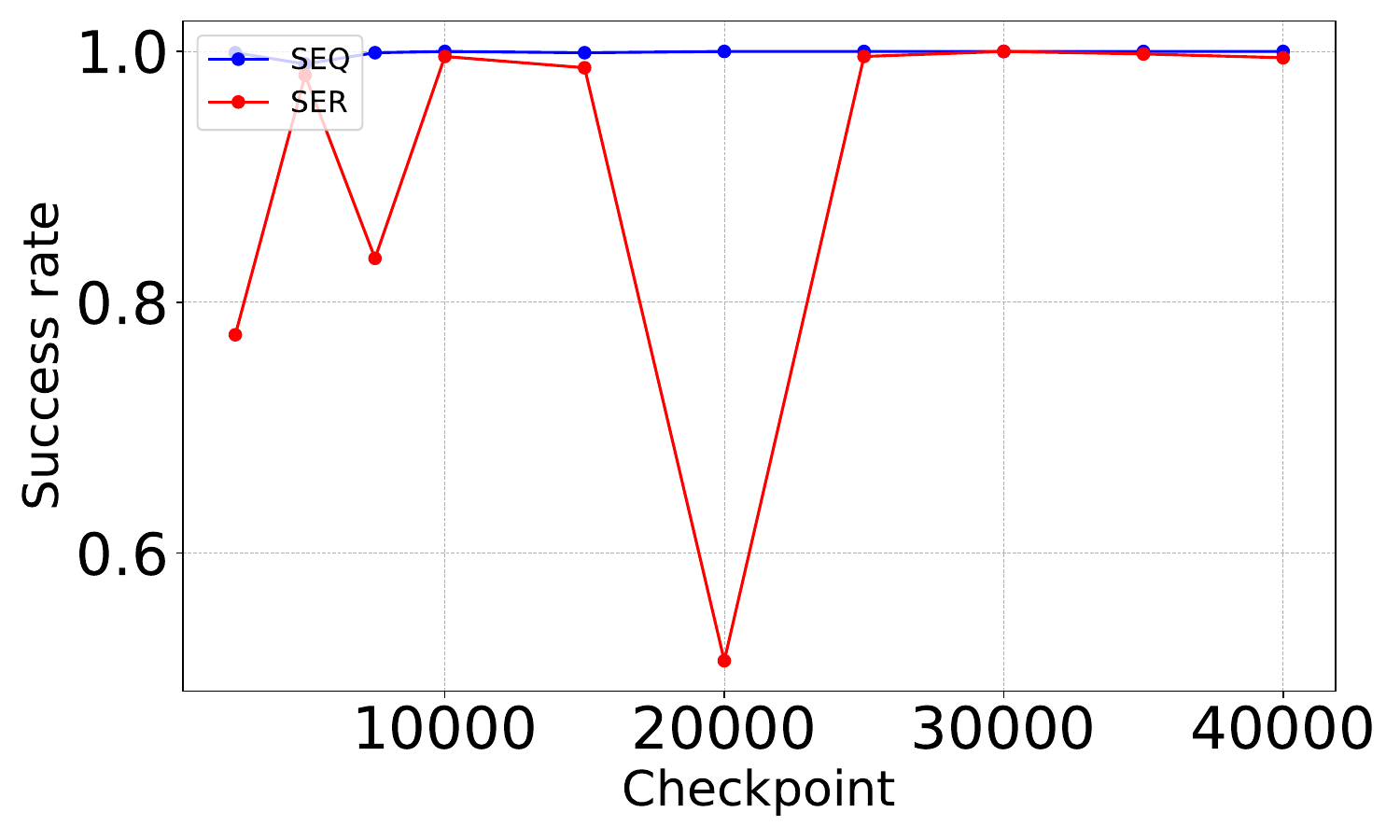}
        \caption{Pre-trained Llama-7b tested on Step 1 }
        \label{fig:llama-check-step}
    \end{subfigure}
    \hfill
    \begin{subfigure}{0.31\linewidth}
        \centering
        \includegraphics[width=\linewidth]{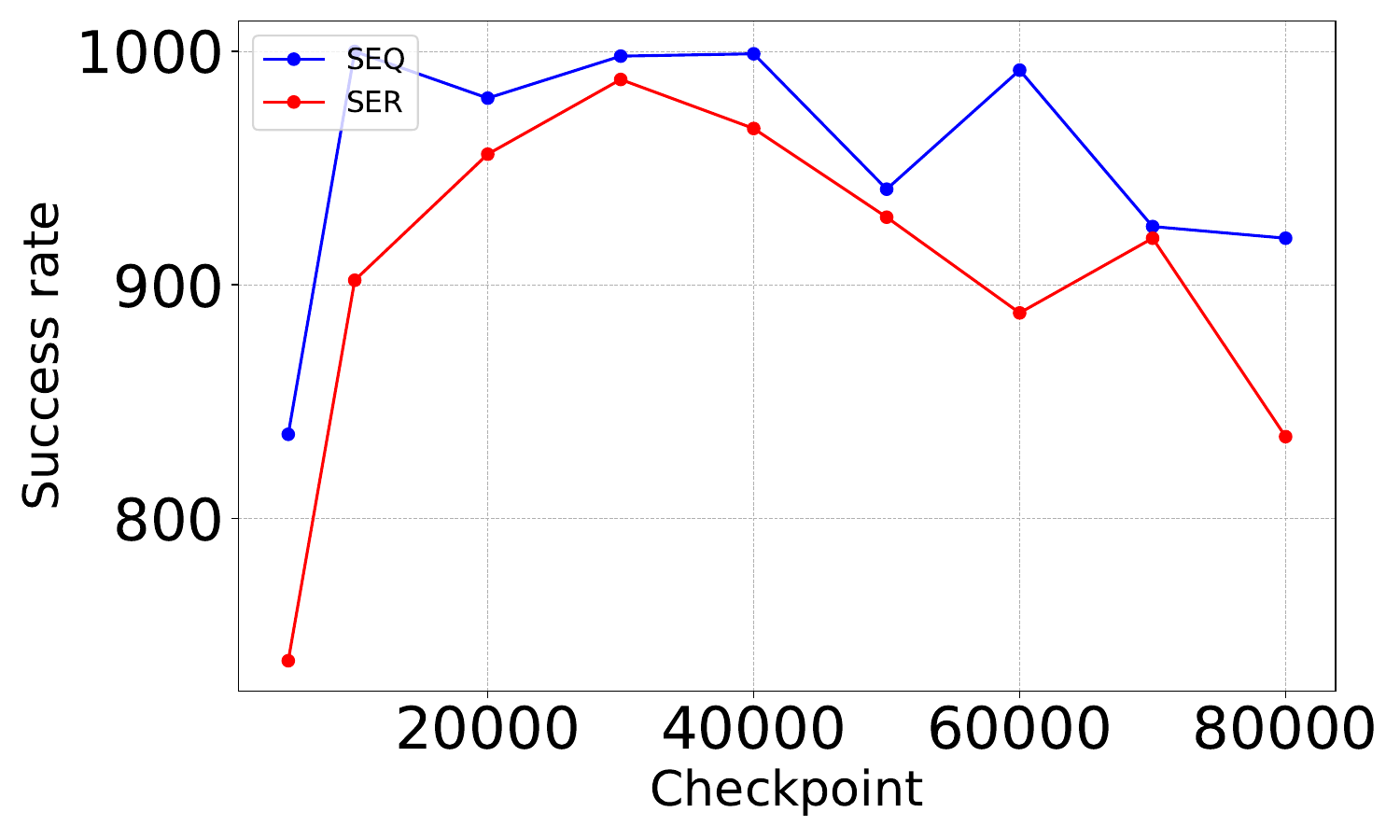}
        \caption{Train-from-scratch Gemma-2b tested on Step 1}
        \label{fig:gemma-train-from-scratch-check-step}
    \end{subfigure}
    \caption{Step 1 success rate for the multiplication task. In (a), (b) and (c), models are prompted with the problem only. Results show that models learn the spurious dependency of Step 1 on Step 2-8, which negatively impact their ability to predict Step 1.}
    \hfill
\end{figure*}

\smallsection{SER models learn spurious dependency}
In this experiment, we test our SEQ and SER models for the multiplication task. The testing prompt of the model is the problem only, and the models are expected to generate the full proof. See Figure~\ref{fig:prompt}a,~\ref{fig:prompt}c. We use the Step 1 correctness metric, i.e. checking the correctness of Step 1 in the models' generated output. Step 1 is the decomposition step that splits the 4-digit number into simple terms, e.g. 2345 to 2*1000+3*100+4*10+5*1. During the training, the model should be able to learn this step quickly given the problem due to its simplicity. Indeed, from Figure~\ref{fig:gemmea-check-step},~\ref{fig:llama-check-step}, we see that for SEQ models based on pre-trained models, they learn Step 1 almost perfectly after the first 2500 steps. For the randomly initialized Gemma-2b trained from scratch on SEQ, the model performance measured by the Step 1 correctness metric quickly reaches more than 90\% and is approaching perfect scores followed by some fluctuations, as shown by Figure~\ref{fig:gemma-train-from-scratch-check-step}. For SER models, however, they converge much slower measured by the Step 1 correctness metric in the early training stage compared to SEQ models based on the evidences shown in Figure~\ref{fig:gemmea-check-step},~\ref{fig:llama-check-step},~\ref{fig:gemma-train-from-scratch-check-step}. This slow convergence in the early training stage shows that SER models try hard to fit the spurious dependency of Step 1 on all other proof steps at first. They need more training steps to realize that Step 1 does not depend on all the other proof steps, including the final answer.

\section{Other Results and Discussion}

\label{discussion}

\smallsection{Bias in pre-training data is not the major cause of the order effect}
One might argue that the order bias in the pre-training dataset could potentially be the primary cause of the order effects observed in this study. However, for the multiplication task, we conduct controlled experiments on both pre-trained models and randomly initialized model, thereby ruling out this possibility given the results as shown in Figure~\ref{fig:gemma-train-from-scratch-final-answer}.

\smallsection{SEQ is better than PSER even when problem and proof steps are masked during training}
In this experiment, we don't use our 9 models mentioned before. We fine-tune pre-trained Gemma-2b on SEQ and PSER but with all tokens except final answers masked out. Therefore the only input-output pair that enters into the loss function is (problem+intermediate proof steps, final answer). During testing, we prompt both models with the problem and the intermediate proof steps. See Figure~\ref{fig:prompt}d,\ref{fig:prompt}e for the prompting configuration. Figure~\ref{fig:gemma-masked} shows that the model fine-tuned on masked SEQ outperforms the model fine-tuned scratch on masked PSER measured by final answer correctness for 
the pre-trained case. 

\smallsection{Models treat the same data with reversed order very differently during training}
Lastly, we study the question of whether the model will converge faster when it is fine-tuned on SEQ after it is fine-tuned on PSER. For the multiplication task, we first fine-tune Gemma-2b on PSER  for 50000 steps, and then keep fine-tuning it on SEQ. As shown in Figure~\ref{fig:gemmea-cot-reversed-then-cot}, we see that the model converges much slower after it's fine-tuned on PSER compared to the case when the pre-trained model is directly fine-tuned on SEQ, measured by the proof correctness metric. This suggests that the knowledge that the model learns from the proof in reverse order doesn't help but hurts it in learning the proof in the right order. Essentially, models treat the same data with reversed order very differently during training.

\smallsection{Discussion: artificiality of the dataset}It is important to note that the datasets used in this study for training and evaluating models are artificially constructed. Ideally, we would prefer to train and evaluate models on graduate- and research-level mathematical theorems and proofs. However, although state-of-the-art (SOTA) models can partially reorder proofs into a more intuitively sequential structure, there remains no reliable method for verifying the correctness of model-generated proofs. To illustrate the widespread nature of ordering issues in mathematical texts, we include in the appendix a list of proofs exhibiting the Dependency-Inversion Issue (DII), drawn from graduate-level mathematics textbooks. These constitute 17.3 percent of the theorems with nontrivial proofs in the first two chapters of the examined textbook, highlighting the severity of ordering problems in advanced mathematical reasoning.

\section{Conclusion and Future work}
In conclusion, our study highlights the significant impact of order for each data sample on LLMs' training efficiency for the proof generation task. We claim that the optimal order of a data sample, which we call the intuitively sequential order, is to put intermediate supervision toward a particular proof step on the left for all proof steps in a data sample. We show the two causes for this order effect: given suboptimal data order, training inefficiency arises from (i) inability to look forward during training, i.e. learning to use future intermediate supervision to predict the current token, (ii) propensity to learn spurious token dependencies in the early training stage. Our experiments on the theorem-proving task and 4-by-4-digit multiplication provide strong support for these explanations, highlighting the importance of finding the optimal data order in model training. We explicitly define a particular kind of order issue, Delayed Insight Issue, that frequently appear in advanced math textbooks, and we present a list of proofs that suffer from DII in the appendix. Notably, 17.3 percent of theorems with non-trivial proofs in the first two chapters of the textbook we analyze suffer from DII. 

A promising direction for future work is to investigate this claim during the pre-training stage of LLMs for reasoning tasks. Another future work is to employ machine learning methods to discover optimal data order. This holds significance on its own, independent of the gain it can bring to model training. For example, suppose the true supervision structures are discovered for mathematical proofs and those proofs are reordered accordingly. In that case, it might facilitate the learning process of students, making it linear and intuitive.

\section{Limitations}
In this paper, we conduct experiments on models with sizes of 2B and 7B. We did not extend our experiments to larger models due to limitations in computational resources.

Furthermore, due to computational constraints, we did not repeat our training-from-scratch experiments multiple times.

\bibliography{iclr2021_conference}

\appendix

\section{Appendix}

\subsection{Limitation}
In this paper, we conduct experiments on models with 2B and 7B parameters. Due to computational constraints, we did not extend our experiments to larger models or incorporate additional ordering variations. As discussed in Section~\ref{discussion}, we didn't train and evaluate models on graduate or research-level math-proving datasets, since there is no direct way to check the correctness of the model-generated proofs. 

\subsection{Delayed Insight Issue: A common type of order issue that occurs frequently in graduate-level math textbooks}
\label{example:folland}

To illustrate DII, we compile a list of 9 cases from the first two chapters of a classic graduate-level textbook: Real Analysis, Modern Techniques and Their Applications authored by Gerald B. Folland \citep{Folland1984RealAM}. We find 17.3 percent of the theorems with non-trival proofs in the first two chapters of the textbook suffer from Delayed Insight Issue as defined in the main text.

\subsubsection{Folland, Chapter 1, Proposition 1.3}
\begin{tcolorbox}[colback=blue!5!white, colframe=blue!75!black, title=Example 1]
Proposition. If $A$ is countable, then $\bigotimes_{\alpha\in A}M_\alpha$ is the $\sigma$-algebra generated by $\{\prod_{\alpha\in A}E_\alpha:E_\alpha\in M_\alpha\}$

\vspace{1em}

Proof: If $E_\alpha\in M_\alpha$, then $\pi_\alpha^{-1}(E_\alpha)=\prod_{\beta\in A}E_\beta$ where $E_\beta=X_\beta$ for $\beta\neq \alpha$; on the other hand, $\prod_{\alpha\in A}E_\alpha = \cap_{\alpha\in A}\pi^{-1}_\alpha(E_\alpha)$. The result therefore follows from Lemma 1.1.
\end{tcolorbox}
Here, Lemma 1.1 serves as the major lemma, and the previous steps set up the stage for applying Proposition 1.4. Note that it is natural to first come up with the idea of applying Lemma 1.1 given the theorem.

\subsubsection{Folland, Chapter 1, Proposition 1.5}
\begin{tcolorbox}[colback=blue!5!white, colframe=blue!75!black, title=Example 2]
Proposition. Let $X_1,..,X_n$ be metric spaces and let $X=\prod_1^n X_j$, equipped with the product metric. Then $\bigotimes_1^n B_{X_j}\subset B_X$. If the $X_j$'s are separable, then $\bigotimes_1^nB_{X_j}=B_X$

\vspace{1em}

Proof: ...... Moreover, the set of the set of points in $X$ whose jth coordinate is in $C_j$ for all j is a countable dense subset of $X$, and the balls of radius $r$ in $X$ are merely products of the balls of radius r in the $X_j$'s. It follows that $B_{X_j}$ is generated by $\mathbb{E}_j$ and $B_X$ is generated by $\prod^n_1E_j:E_j\in\mathbb{E}_j$. Therefore $B_X=\bigotimes_1^nB_{X_j}$ by Proposition 1.4
\end{tcolorbox}
Here, Proposition 1.4 serves as the major lemma, and the previous steps set up the stage for applying Proposition 1.4. Note that it is natural to first come up with the idea of applying Proposition 1.4 given the theorem.

\subsubsection{Folland, Chapter 1, Proposition 1.16}
\begin{tcolorbox}[colback=blue!5!white, colframe=blue!75!black, title=Example 3]
Proposition. If $F:\mathbb{R}\mapsto\mathbb{R}$ is any increasing, right continuous function, there is a unique Borel measure $\mu_F$
 on $\mathbb{R}$ such that $\mu_F((a,b])=F(b)-F(a)$ for all $a,b$. If $G$ is another such function, we have $\mu_F = \mu_G$ iff $F-G$ is a constant.
\vspace{1em}

Proof: Each $F$ induces a premeasure on $A$ by Proposition 1.15. It is clear that $F$ and $G$ induces the same premeasure iff $F-G$ is constant, and that these premeasures are $\sigma$-finite (since $\mathbb{R} = \cup^\infty_{-\infty}(j,j+1]$). The first two assertions therefore follow from Theorem 1.14.
\end{tcolorbox}
Here, Theorem 1.14 serves as the major lemma, and the previous steps set up the stage for applying Theorem 1.14. Note that it is natural to first come up with the idea of applying Theorem 1.14 given the theorem.

\subsubsection{Folland, Chapter 2, Proposition 2.6}
\begin{tcolorbox}[colback=blue!5!white, colframe=blue!75!black, title=Example 4]
Proposition. If $f,g:X\mapsto \mathbb{C}$ are $M$-measurable, then so are $f+g$ and $fg$
\vspace{1em}

Proof: Define $F:X\mapsto \mathbb{C}\times\mathbb{C}$, $\phi:\mathbb{C}\times\mathbb{C}\mapsto \mathbb{C}$ by $F(x)=(f(x),g(x))$, $\phi(z,w)=z+w$, $\psi(z,w)=zw$. Since $B_{\mathbb{C}\times\mathbb{C}}=B_{\mathbb{C}}\bigotimes B_{\mathbb{C}}$ by Proposition 1.5, $F$ is $(M, B_{\mathbb{C}\times\mathbb{C}})$-measurable by Proposition 2.4.
\end{tcolorbox}
Here, Proposition 2.4 serves as the major lemma, and the previous steps set up the stage for applying Theorem 1.14. Note that it is natural to first come up with the idea of applying Proposition 2.4 given the theorem, and then make the constructions as in the early part of the proof.

\subsubsection{Folland, Chapter 2, Corollary 2.17}
\begin{tcolorbox}[colback=blue!5!white, colframe=blue!75!black, title=Example 5]
Corollary. If $\{f_n\}\subset L^+$, $f\in L^+$, and $f_n(x)$ increases to $f(x)$ for a.e $x$, then $\int f = \lim\int f_n$
\vspace{1em}

Proof: If $f_n(x)$ increases to $f(x)$ for $x\in E$ where $\mu(E^c) = 0$, then $f-f\chi_E=0$ a.e. and $f_n-f_n\chi_E=0$ a.e., so by the monotone convergence theorem, $\int f=\int f\chi_E=\lim\int f_n\chi_E=\lim\int f_n$ 
\end{tcolorbox}
For this problem, the first instinct after looking at the corollary should be to apply the monotone convergence theorem. The earlier steps explicitly state how the monotone convergence theorem should be applied to the given corollary.

\subsubsection{Folland, Chapter 2, Fatou's Lemma 2.18}
\begin{tcolorbox}[colback=blue!5!white, colframe=blue!75!black, title=Example 6]
Fatou's Lemma. If $\{f_n\}$ is any sequence in $L^+$, then $\int(\lim \inf f_n)\leq \lim\inf\int f_n$
\vspace{1em}

Proof: For each $k\geq 1$ we have $\inf_{n\geq k}f_n\leq f_j$ for $j\geq k$, hence $\int\inf_{n\geq k}f_n\leq \int f_j$ for $j\geq k$, hence $\int\inf_{n\geq k}f_n\leq\inf_{j\geq k}\int f_j$. Now let $k\mapsto \infty$ and apply monotone convergence theorem: $\int(\lim\inf f_n)=\lim_{k\mapsto\infty}\int(\inf_{n\geq k}f_n)\leq \lim\inf\int f_n$
\end{tcolorbox}
For this problem, the first instinct after looking at the lemma should be to apply the monotone convergence theorem. The earlier steps explicitly state how to transform the problem to a state so that one can apply the monotone convergence theorem.

\subsubsection{Folland, Chapter 2, Theorem 2.25}
\begin{tcolorbox}[colback=blue!5!white, colframe=blue!75!black, title=Example 7]
Theorem. Suppose that $\{f_j\}$ is a sequence in $L^1$ such that $\Sigma_1^\infty\int|f_j|<\infty$. Then $\Sigma_1^\infty f_j$ converges a.e. to a function in $L^1$, and $\int\Sigma_1^\infty f_j=\Sigma_1^\infty\int f_j$
\vspace{1em}

Proof: By Theorem 2.15, $\int\Sigma_1^\infty|f_j|=\Sigma_1^\infty\int|f_j|<\infty$, so that function $g= \Sigma_1^\infty|f_j|$ is in $L^1$. In particular, by Proposition 2.20, $\Sigma_1^\infty|f_j(x)|$ is finite for a.e. $x$, and for each such $x$ the series $\Sigma_1^\infty f_j(x)$ converges. Moreover, $|\Sigma_1^n f_j|\leq g$ for all $n$, so we can apply the dominated convergence theorem to the sequence of partial sums to obtain $\int\Sigma_1^\infty f_j = \Sigma_1^\infty\int f_j$
\end{tcolorbox}
For this problem, the first instinct after looking at the lemma should be to apply the dominated convergence theorem. The earlier steps explicitly state how to make sure the conditions of the dominated convergence theorem are satisfied by the statement.

\subsubsection{Folland, Chapter 2, Theorem 2.27}
\begin{tcolorbox}[colback=blue!5!white, colframe=blue!75!black, title=Example 8]
Theorem. Suppose that $f:X\times [a,b]\mapsto \mathbb{C} (-\infty<a<b<\infty)$ and that $f(\cdot, t):X\mapsto \mathbb{C}$ is integrable for each $t\in [a,b]$. Let $F(t)=\int_X f(x,t)d\mu(x)$. Suppose that $\partial f/\partial t$ exists and there is a $g\in L^1(\mu)$ such that $|(\partial f/\partial t)(x,t)|\leq g(x)$ for all $x,t$. Then $F$ is differentiable and $F'(x)=\int(\partial f/\partial t)(x,t)d\mu(x)$.  
\vspace{1em}

Proof: Observe that $\partial f/\partial t(x,t_0)=\lim h_n(x)$ where $h_n(x)=\frac{f(x,t_n)-f(x,t_0)}{t_n-t_0}$, $\{t_n\}$ again being any sequence converging to $t_0$. It follows that $\partial f/\partial t$ is measurable, and by the mean value theorem, $|h_n(x)|\leq \sup_{t\in[a,b]}|\partial f/\partial t(x,t)|\leq g(x)$, so the dominated convergence theorem can be invoked again to give $F'(t_0)=\lim\frac{F(t_n)-F(t_0)}{t_n-t_0}=\lim\int h_n(x)d\mu(x)=\int \partial f/\partial t(x,t)d\mu(x)$
\end{tcolorbox}
For this problem, the first instinct after looking at the lemma should be to apply the dominated convergence theorem. The earlier steps explicitly state how to make sure the conditions of the dominated convergence theorem are satisfied by the statement.

\subsubsection{Folland, Chapter 2, Theorem 2.47}

\begin{tcolorbox}[colback=blue!5!white, colframe=blue!75!black, title=Example 9]
Theorem. Suppose that $\Omega$ is an open set in $\mathbb{R}^n$ and $G:\Omega\mapsto \mathbb{R}^n$ is a $C^1$ diffeomorphism. If $f$ is a Lebesgue measurable function on $G(\Omega)$, then $f\circ G$ is Lebesgue measurable on $\Omega$. If $f\geq 0$ or $f\in L^1(G(\Omega),m)$, then $\int_{G(\Omega)} f(x)dx=\int_{\Omega}f\circ G(x)|\det D_xG|dx$.

\vspace{1em}

Proof: ...A lot of estimation....By the preceding estimate and the dominated convergence theorem, $m(G(E))\leq m(G(\cap^\infty_1 U_j))=\lim m(G(U_j))\leq \lim\int_{U_j}|\det D_xG|dx = \int_E|\det D_xG|dx$. Finally, if $E$ is any Borel subset...
\end{tcolorbox}
For this problem, without first having a clear goal as expressed in the previous equation, one will be completely confused for why we need to do all the estimation at the beginning.

\subsection{Prompting configuration for testing}
\begin{figure*}[t]
\label{prompt}
    \centering        \includegraphics[width=0.8\linewidth]{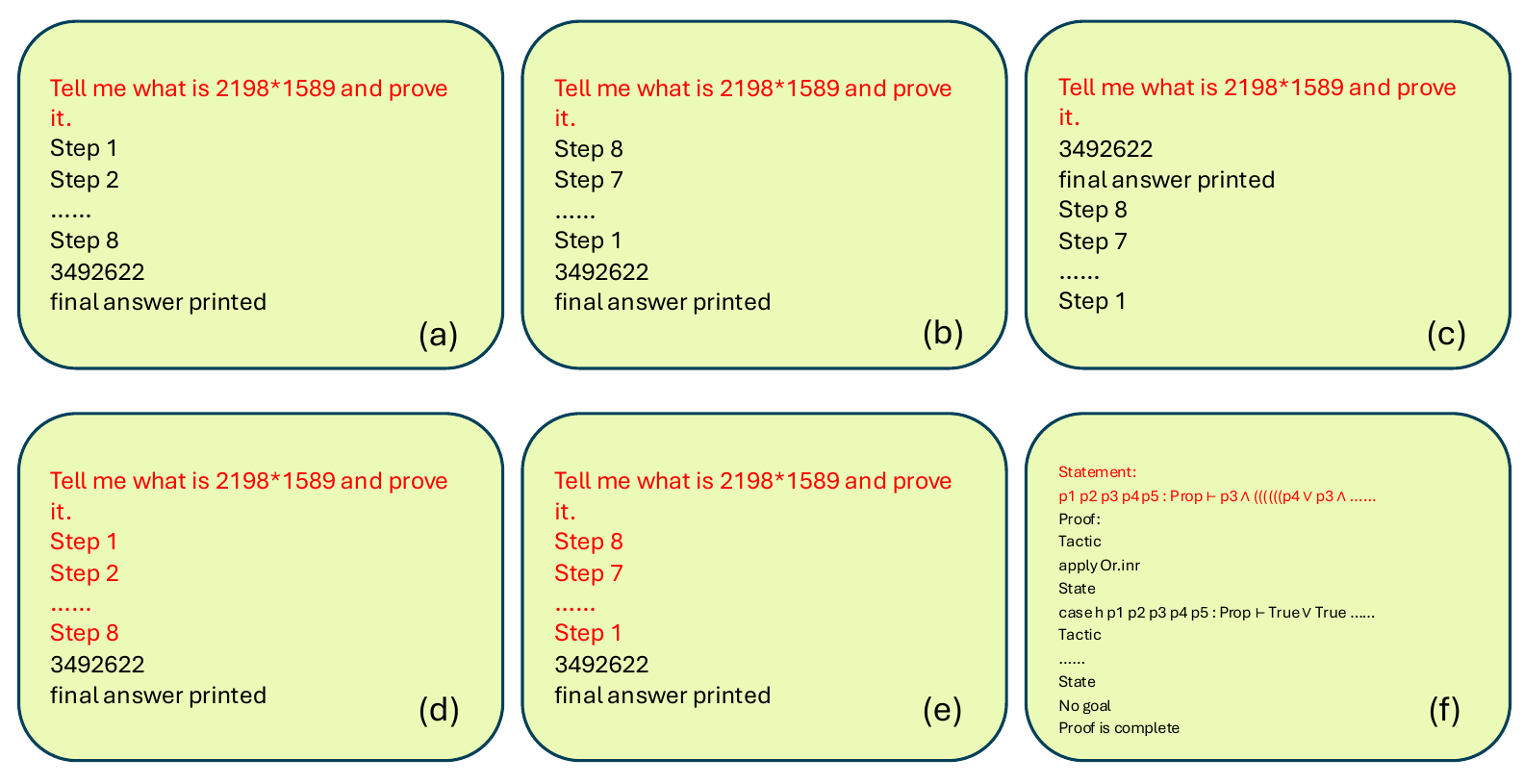}
        \caption{Prompt configuration for testing the models on the multiplication task and the theorem-proving task: texts colored in red are prompt. Texts colored in black are the expected output from the model. }
        \label{fig:prompt}
\end{figure*}

See Figure~\ref{fig:prompt}.

\subsection{Data Examples in training}
\begin{figure}[t]
    \centering
    \includegraphics[width=\linewidth]{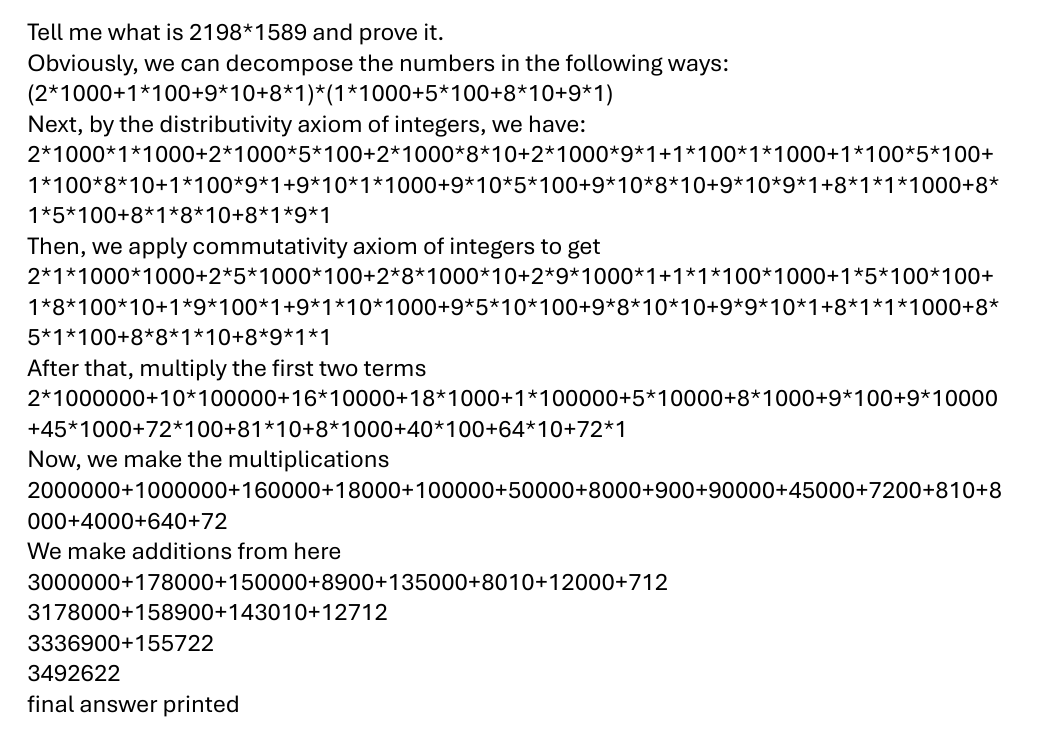}
    \caption{Example of the SEQ data for multiplication}
    \label{fig:data-example-seq}
\end{figure}

\begin{figure}[t]
    \centering
    \includegraphics[width=\linewidth]{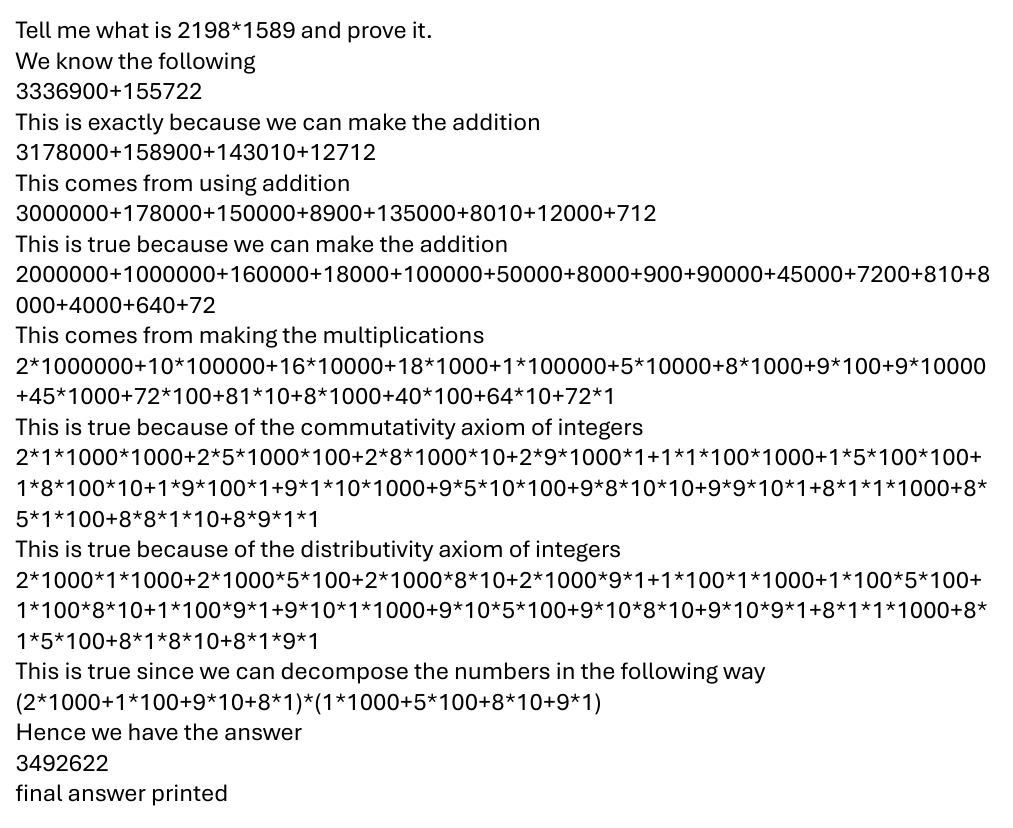}
    \caption{Example of the PSER data for multiplication}
    \label{fig:data-example-pser}
\end{figure}

\begin{figure}[t]
    \centering
    \includegraphics[width=\linewidth]{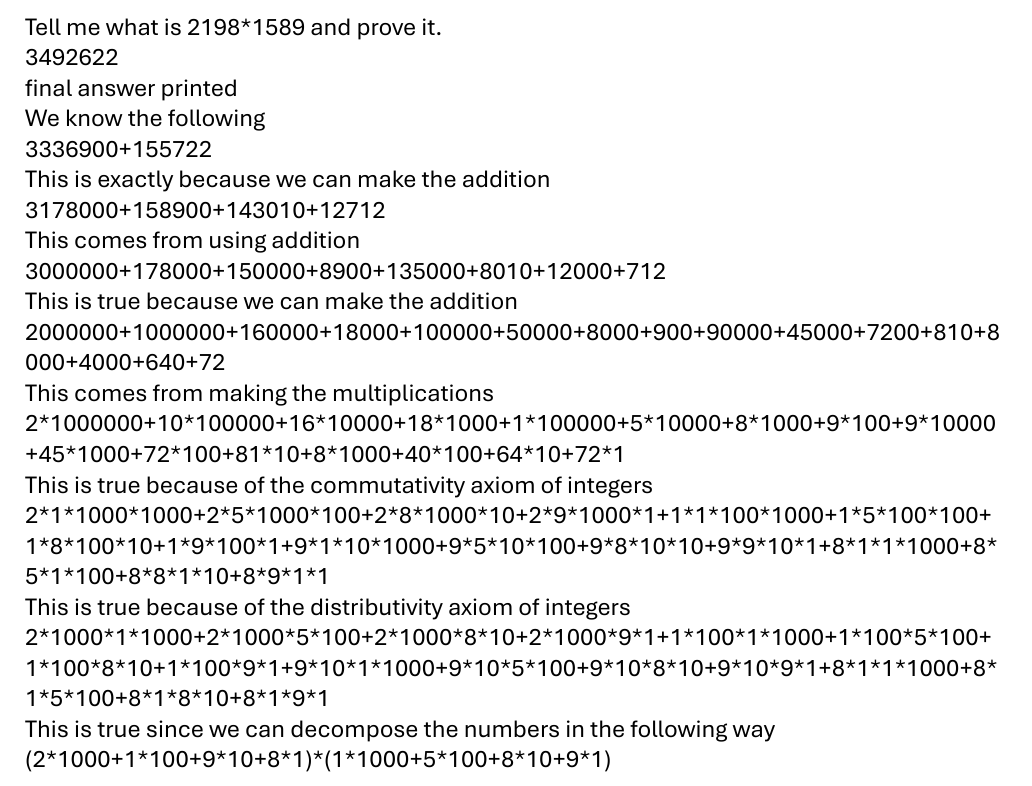}
    \caption{Example of the SER data for multiplication}
    \label{fig:data-example-ser}
\end{figure}

\begin{figure*}[t]
    \centering
\begin{subfigure}{0.4\linewidth}
        \centering
        \includegraphics[width=\linewidth]{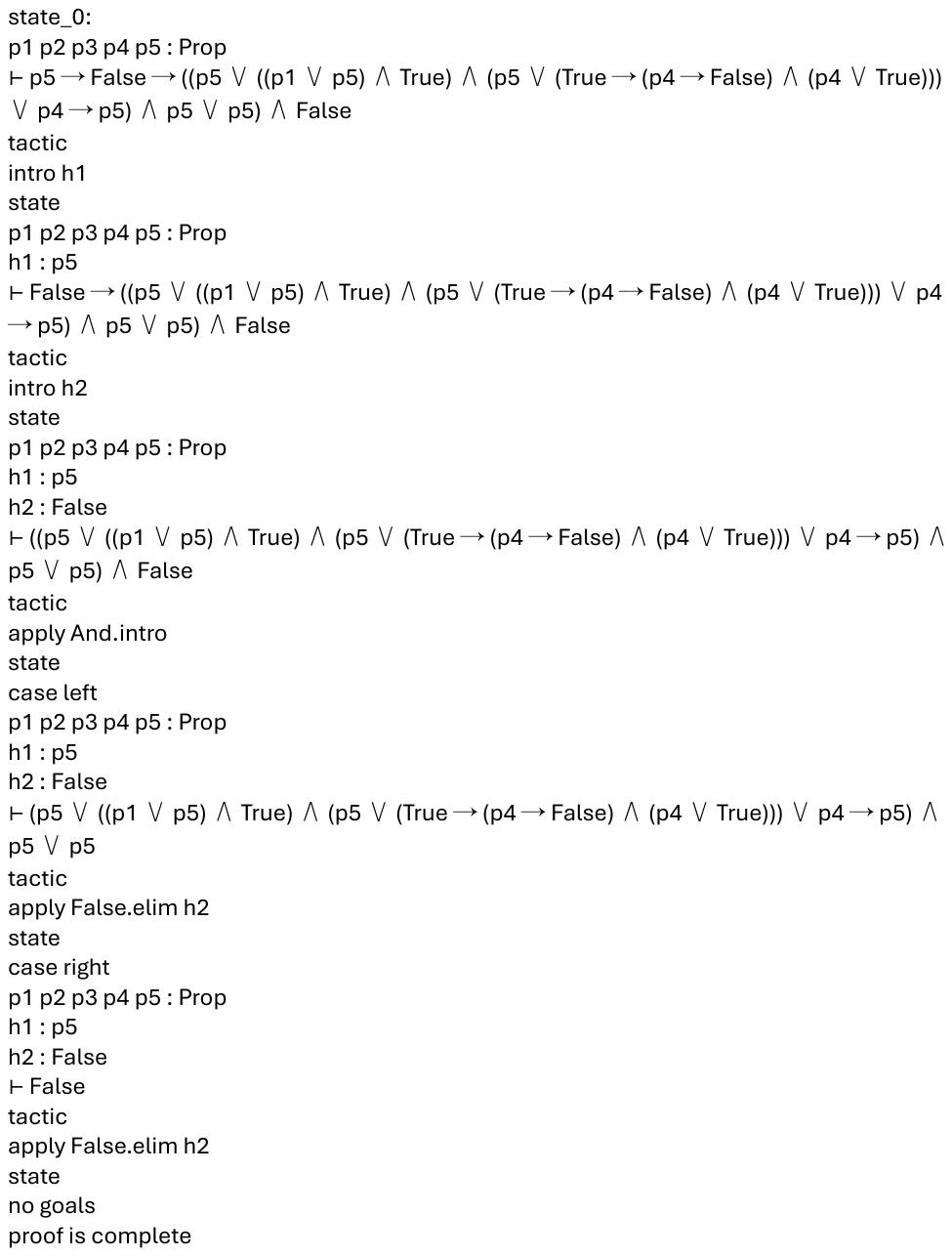}
        \caption{Example of the SEQ data for theorem-proving}
    \end{subfigure}
\hfill
\begin{subfigure}{0.4\linewidth}
        \centering
        \includegraphics[width=\linewidth]{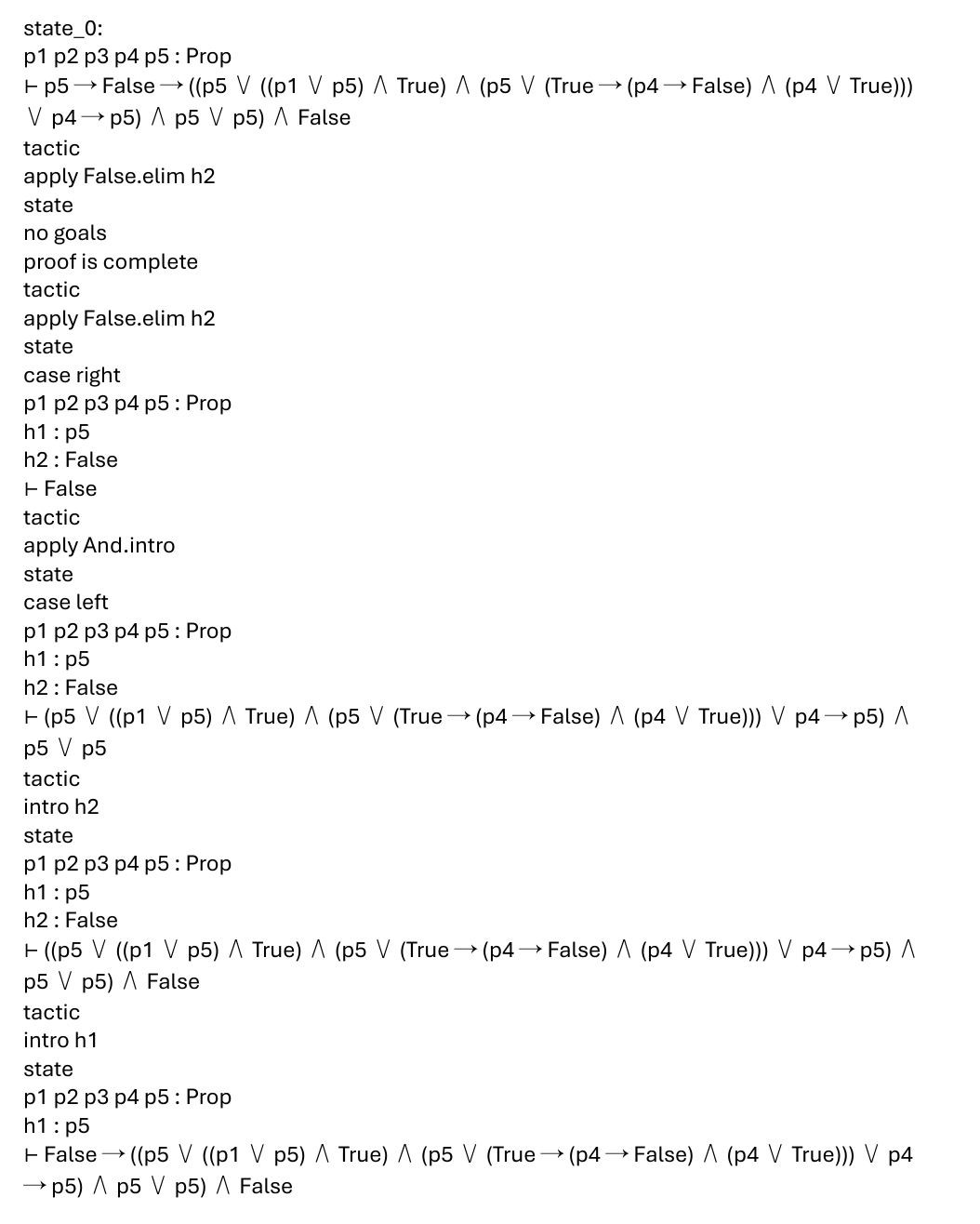}
        \caption{Example of the SER data for theorem-proving}
    \end{subfigure}

    \caption{Examples of data for the intuitionistic propositional logic theorem-proving task}
    \label{fig:data-example-propl}

\end{figure*}

We show here examples of our data for the multiplication task and the intuitionistic propositional logic theorem-proving task. See Figure~\ref{fig:data-example-seq},~\ref{fig:data-example-pser},~\ref{fig:data-example-ser},~\ref{fig:data-example-propl}.
\label{appendix:data-example}

\subsection{Fine-grained result regarding data order}
\begin{figure*}[t]
    \centering        
    \begin{subfigure}{0.5\linewidth}
        \centering        \includegraphics[width=\linewidth]{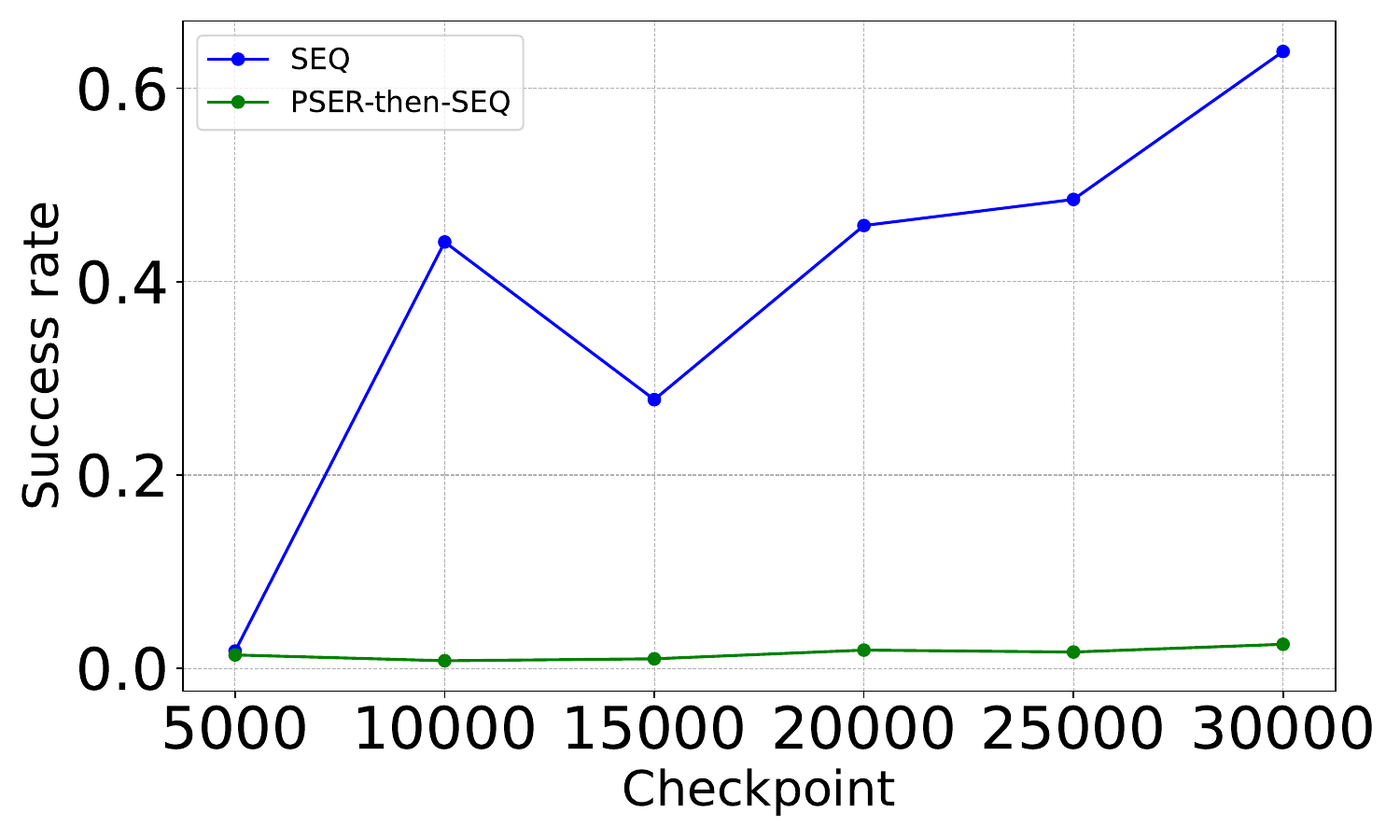}
        \caption{Gemma first fine-tuned on PSER then fine-tuned on SEQ}
        \label{fig:gemmea-cot-reversed-then-cot}
    \end{subfigure}
    \hspace{0.8cm}
    \begin{subfigure}{0.5\linewidth}
        \centering
        \includegraphics[width=\linewidth]{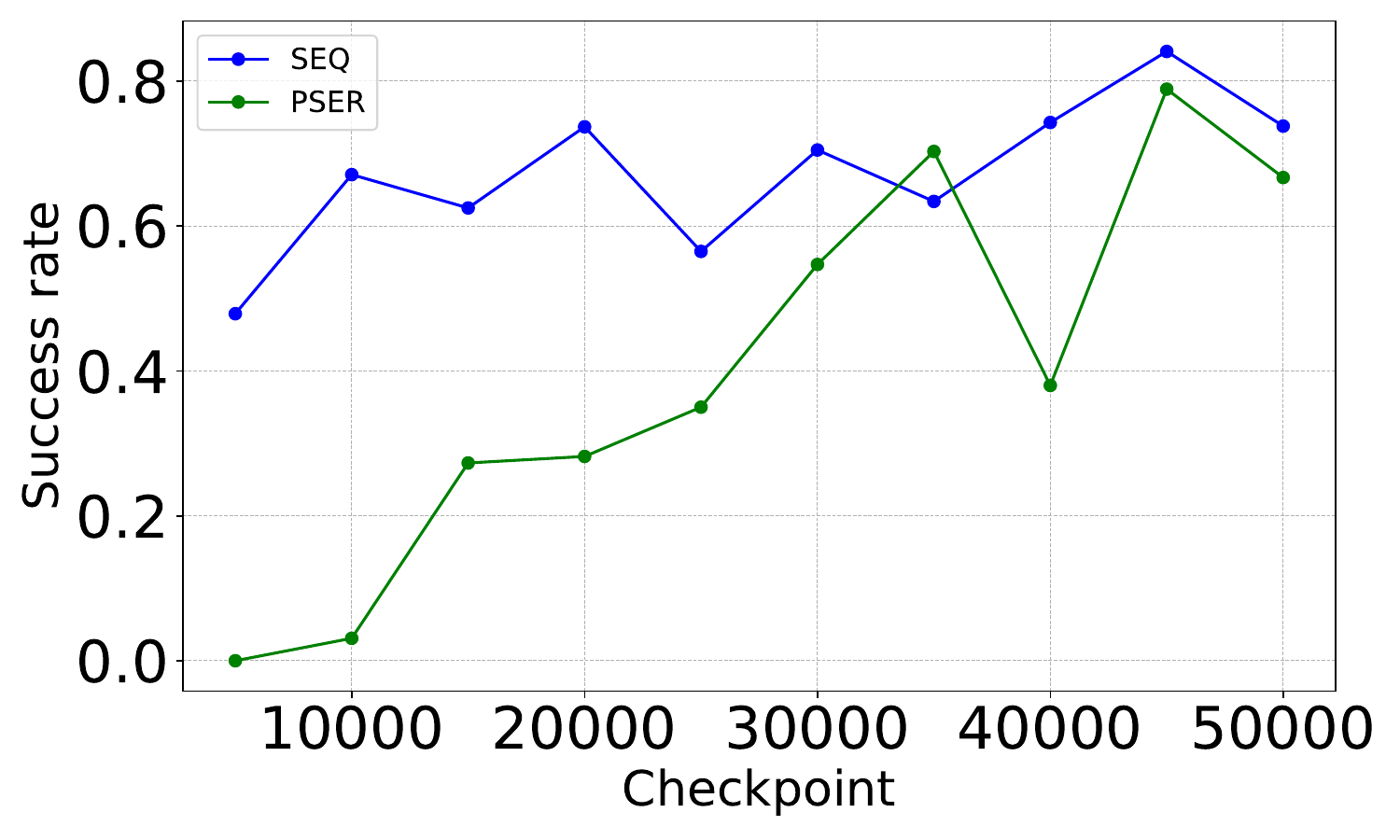}
        \caption{Pre-trained Llama-2-7b-hf, problems and steps masked}
        \label{fig:gemma-masked}
    \end{subfigure}
    \caption{More fine-grained result on the order effect in the multiplication task}
    \label{fig:fine-grained-result}
\end{figure*}

In Figure~\ref{fig:fine-grained-result} we present more fine-grained results on the order effect in the multiplication task.

\end{document}